\documentclass[twoside,11pt]{article}

\usepackage{jmlrwcp2e}

\usepackage{graphicx}
\usepackage{latexsym}
\usepackage{epsf}

\usepackage{bbm,amsmath,amsfonts}
\usepackage[all]{xy}
\usepackage{graphicx}
\usepackage{algorithm,algorithmic}
\usepackage{url}
\usepackage{psfrag}

\usepackage{amsbsy}

\newcommand{\argmax}{\mathop{\mathrm{argmax}}}
\newcommand{\argmin}{\mathop{\mathrm{argmin}}}

\newcommand{\bphi}{\boldsymbol \phi}
\newcommand{\bw}{\mathbf w}
\newcommand{\bhatw}{\mathbf{\hat w}}
\newcommand{\bx}{\mathbf x}

\newcommand{\bd}{\mathbf d}
\newcommand{\btx}{\mathbf{\tilde x}}
\newcommand{\bK}{\mathbf K}

\newcommand{\bI}{\mathbf I}
\newcommand{\bP}{\mathbf P}
\newcommand{\by}{\mathbf y}
\newcommand{\bkm}[1]{\mathbf k_m(#1)}

\newcommand{\bLambda}{\boldsymbol \Lambda}

\newcommand{\definedas}{:=}

\newcommand{\trans}{^\text{\sffamily T}} 
\newcommand{\norm}[1]{\left\|{#1}\right\|}      
\newcommand{\rowvector}[1]{\bigl(#1\bigr)\trans}

\newcommand {\bKmmi}{\mathbf K_{mm}^{-1}}
\newcommand {\bKmm}{\mathbf K_{mm}}

\newcommand{\BV}{\mathcal{BV}}

\newcommand{\bB}{\mathbf B}

\newcommand{\rtt}{r_{t+1}}
\newcommand{\bxtt}{\mathbf x_{t+1}}
\newcommand{\bxt}{\mathbf x_{t}}

\newcommand{\bat}{\mathbf a_t}
\newcommand{\batt}{\mathbf a_{t+1}}

\newcommand{\bPitm}{\mathbf P_{tm}^{-1}}
\newcommand{\bPittm}{\mathbf P_{t+1,m}^{-1}}
\newcommand{\bPittmm}{\mathbf P_{t+1,m+1}^{-1}}

\newcommand{\bPttm}{\mathbf P_{t+1,m}}
\newcommand{\bPttmm}{\mathbf P_{t+1,m+1}}

\newcommand{\bPtm}{\mathbf P_{tm}}

\newcommand{\xitm}{\xi_{tm}}
\newcommand{\xittm}{\xi_{t+1,m}}
\newcommand{\xittmm}{\xi_{t+1,m+1}}

\newcommand{\bkt}{\mathbf k_{t}}
\newcommand{\bktt}{\mathbf k_{t+1}}

\newcommand{\bhtt}{\mathbf h_{t+1}}

\newcommand{\bztm}{\mathbf z_{tm}}
\newcommand{\bzttm}{\mathbf z_{t+1,m}}
\newcommand{\bzttmm}{\mathbf z_{t+1,m+1}}

\newcommand{\kstar}{k^{*}}

\newcommand{\hstar}{h^{*}}

\newcommand{\bq}{\mathbf q}

\newcommand{\bzeros}{\mathbf 0}
\newcommand{\bwb}{\mathbf w_b}

\newcommand{\bb}{\mathbf b}
\newcommand{\bstar}{b^*}

\newcommand{\Jtm}{J_{tm}}
\newcommand{\brt}{\mathbf r_t}
\newcommand{\brtt}{\mathbf r_{t+1}}

\newcommand{\bwtm}{\bw_{tm}}

\newcommand{\bwttm}{\bw_{t+1,m}}
\newcommand{\bwttmm}{\bw_{t+1,m+1}}

\newcommand{\bbtm}{\mathbf b_{tm}}
\newcommand{\bbttm}{\mathbf b_{t+1,m}}
\newcommand{\bbttmm}{\mathbf b_{t+1,m+1}}

\newcommand{\bAtm}{\mathbf A_{tm}}
\newcommand{\bAttm}{\mathbf A_{t+1,m}}
\newcommand{\bAttmm}{\mathbf A_{t+1,m+1}}

\newcommand{\bKtm}{\mathbf K_{tm}}
\newcommand{\bKttm}{\mathbf K_{t+1,m}}

\newcommand{\bHtm}{\mathbf H_{tm}}
\newcommand{\bHttm}{\mathbf H_{t+1,m}}
\newcommand{\bHttmm}{\mathbf H_{t+1,m+1}}

\newcommand{\bZttm}{\mathbf Z_{t+1,m}}

\jmlrheading{1}{2007}{33-57}{}{}{Tobias Jung and Daniel Polani}{Gaussian Processes in Practice}

\ShortHeadings{Learning Keepaway with Kernels}{Jung and Polani}
\firstpageno{33}

\begin{document}

\title{Learning RoboCup-Keepaway with Kernels}

\author{\name Tobias Jung \email tjung@informatik.uni-mainz.de \\
       \addr Department of Computer Science\\
       University of Mainz\\
       55099 Mainz, Germany
       \AND
       \name Daniel Polani \email d.polani@herts.ac.uk \\
       \addr School of Computer Science \\
       University of Hertfordshire\\
       Hatfield AL10 9AB, UK}

\editor{Neil D. Lawrence, Anton Schwaighofer and Joaquin Qui\~nonero Candela}

\maketitle

\begin{abstract}
We apply kernel-based methods to solve the difficult reinforcement learning
problem of 3vs2 keepaway in RoboCup simulated soccer. Key challenges in keepaway 
are the high-dimensionality of the state space (rendering conventional
discretization-based function approximation like tilecoding infeasible), 
the stochasticity due to noise and multiple learning agents needing 
to cooperate (meaning that the exact dynamics of the environment are
unknown) and real-time learning (meaning that an efficient online implementation
is required). We employ the general framework of approximate policy iteration
with least-squares-based policy evaluation.
As underlying function approximator we consider the family of regularization
networks with subset of regressors approximation. The core of our proposed
solution is an efficient recursive implementation with automatic supervised
selection of relevant basis functions. 
Simulation results indicate that the behavior learned through our approach 
clearly outperforms the best results obtained with tilecoding 
by \citet{AB05}.\end{abstract}


\section{Introduction}
RoboCup simulated soccer has been conceived and is widely accepted as a 
common platform to address various challenges in artificial intelligence and
robotics research. 
Here, we consider a subtask of the full problem, namely the {\em keepaway} problem. In  
{\em keepaway} we have two smaller teams: one team (the `keepers') must try to maintain 
possession of the ball for as long as possible while staying within a small region of the full soccer field. 
The other team (the `takers') tries to gain possession of the ball. 
\citet{AB05} initially formulated keepaway as benchmark problem for reinforcement learning (RL); 
the keepers must individually {\em learn} how to maximize the time they control the ball as a team 
against the team of opposing takers playing a fixed strategy. 
The central challenges to overcome are, for one, the high 
dimensionality of the state space (each observed state is a vector of 13 measurements), meaning 
that conventional approaches to function approximation in RL, like grid-based tilecoding, are infeasible; 
second, the 
stochasticity due to noise and the uncertainty in control due to the multi-agent nature imply that
the dynamics of the environment are both unknown and cannot be obtained easily. Hence we need 
model-free methods. Finally, the underlying soccer server expects an action every 100 msec,
meaning that efficient methods are necessary that are able to learn in real-time.

\citet{AB05} successfully applied RL to {\em keepaway}, using the 
textbook approach with online Sarsa($\lambda$) and 
tilecoding as underlying function approximator \citep{sutton98introduction}. However, tilecoding is 
a local method and  places parameters (i.e.\ basis functions) in a regular fashion throughout the 
entire state space, 
such that the number of parameters grows exponentially with the dimensionality of the space. In \citep{AB05} this very 
serious shortcoming
was adressed by exploiting  
problem-specific knowledge of how the various state variables interact. In particular, each state variable 
was considered independently from the rest. Here, we will demonstrate that one can also learn using the full 
(untampered) state information, without resorting to simplifying assumptions. 

In this paper we propose a (non-parametric) kernel-based approach to approximate the value function. 
The rationale for doing this is that by representing the solution through the data and not by some
basis functions chosen before the data becomes available, we can better adapt to the complexity of 
the unknown function we are trying to estimate. In particular, parameters 
are not `wasted' on parts of the input space that are never visited. The hope is that thereby the
exponential growth of parameters 
is bypassed. To solve the RL problem
of optimal control we consider the framework of approximate policy iteration with the related
least-squares based
policy evaluation methods LSPE($\lambda$) proposed by \citet{nedicbert2003LSPE} and LSTD($\lambda$) proposed by \citet{boyan99lstd}. Least-squares based policy evaluation is
ideally suited for the use with linear models and is a very sample-efficient variant of RL. In this paper 
we provide a unified and concise formulation of LSPE and LSTD; the 
approximated value function is obtained from a regularization network 
which is 
effectively the mean of the posterior obtained by GP regression \citep{raswil06gpbook}. We use the subset of regressors method \citep{smola2000SGMA,luowahba97has} to approximate 
the kernel using a much reduced subset of basis functions.
To select this subset we employ greedy online selection, similar to  
\citep{csato2001sparse,engel2003gptd}, that adds a 
candidate basis function based on its distance to the span of the previously chosen ones. One improvement is
that we consider a {\em supervised} criterion for the selection of the relevant basis functions 
that takes into account the reduction of the cost in the original learning task 
in addition to reducing the error incurred from 
approximating the kernel. Since the per-step complexity during training and prediction depends 
on the size of the subset, making sure that no unnecessary basis functions are selected ensures 
more efficient usage of otherwise scarce resources. In this way learning in real-time 
(a necessity for {\em keepaway}) becomes possible.  

This paper is structured in three parts: the first part (Section~\ref{sec:background}) 
gives a brief introduction on reinforcement learning and carrying out general regression with regularization networks. 
The second part (Section~\ref{sec:pe with rn}) describes and derives an efficient recursive implementation of the proposed
approach, particularly suited for online learning. The third
part describes the RoboCup-keepaway problem in more detail (Section~\ref{sec:robocup}) and contains
the results we were able to achieve (Section~\ref{sec:experiments and results}). 
A longer discussion of related work is deferred to the end of the paper; there we 
compare the similarities of our work with that of \citet{engel2003gptd,engel2005rlgptd,engel2005octopus}.

\section{Background}
\label{sec:background}
In this section we briefly review the subjects of RL and regularization networks.

\subsection{Reinforcement Learning}
%
Reinforcement learning (RL) is a simulation-based form of approximate dynamic programming, e.g. see 
\citep{bert96neurodynamicprogram}. Consider a 
discrete-time dynamical system with states $\mathcal S=\{1,\ldots,N\}$ (for ease of exposition we
assume the finite case). At each time step $t$, when the system is in state $s_t$, a decision
maker chooses a control-action $a_t$ (again, selected from a finite set $\mathcal A$ of admissible actions)
which changes probabilistically the state of the system to $s_{t+1}$, with distribution $P(s_{t+1}|s_t,a_t)$. 
Every such transition yields an immediate reward $r_{t+1}=R(s_{t+1}|s_t,a_t)$. The ultimate goal
of the decision-maker is to choose a course of actions such that the long-term performance, a measure of the
cumulated sum of rewards, is maximized.

\subsubsection{Model-free Q-value function and optimal control}
%
Let $\pi$ denote a decision-rule (called the policy) that maps states to actions. For a fixed
policy $\pi$ we want to evaluate the state-action value function (Q-function) which for every state $s$ is
taken to be the expected infinite-horizon discounted sum of rewards obtained from starting in state $s$,
choosing action $a$ and then proceeding to select actions according to $\pi$:
\begin{equation}
\label{eq: Definition von Q}
Q^\pi(s,a)\definedas E^\pi \left\{ \sum_{t\ge0} \gamma^t r_{t+1} |s_0=s, a_0=a \right\} \quad \forall s,a 
\end{equation}
where $s_{t+1} \sim P(\cdot \ |s_t,\pi(s_t))$ and $r_{t+1}=R(s_{t+1}|s_t,\pi(s_t))$. The parameter 
$\gamma\in (0,1)$ denotes a discount factor.

Ultimately, we are not directly interested in $Q^\pi$; our true goal is optimal control, i.e.\ we seek an 
optimal policy $\pi^*=\argmax_\pi Q^\pi$. To accomplish
that, policy iteration interleaves the two steps policy evaluation and policy improvement: 
First, compute
$Q^{\pi_k}$ for a fixed policy $\pi_k$. Then, once $Q^{\pi_k}$ is known, derive an improved policy $\pi_{k+1}$
by choosing the greedy policy with respect to $Q^{\pi_k}$, i.e. by  
by choosing in every state the action $\pi_{k+1}(s)=\argmax_a Q^{\pi_k}(s,a)$
that achieves the best Q-value. Obtaining the best action is trivial if we employ the
Q-notation, otherwise we would need the transition probabilities and reward function (i.e.\ a `model').

To compute the Q-function, one exploits the fact that $Q^\pi$ obeys the fixed-point relation $Q^\pi=\mathcal T_\pi Q^\pi$,
where $\mathcal T_\pi$ is the Bellman operator
\[
\bigl(\mathcal T_\pi Q\bigr)(s,a)\definedas E_{s'\sim P(\cdot \ |s,a)} \left\{R(s'|s,a)+\gamma Q(s',\pi(s')) \right\}.
\] 
In principle, it is possible to calculate $Q^\pi$
exactly by solving the corresponding linear system of equations, provided that the transition probabilities $P(s'|s,a)$
and rewards $R(s'|s,a)$ are known in advance and the number of states is finite and small.

\begin{figure}
\psfrag{pik}{{\small $\pi_k$}}
\psfrag{pikk}{{\small $\pi_{k+1}$}}
\psfrag{wk}{{\small $\bw_{k}$}}
\psfrag{Q}{{\tiny $\tilde Q(\cdot \ ;\bw_k) \approx Q^{\pi_k}$}}
\psfrag{max}{{\tiny  $\tilde Q(\cdot \ ;\bw_k)$}}
\psfrag{s0a0}{{\tiny  $\{s_i,a_i,r_{i+1},s_{i+1}\}$}}
	\centering
	\includegraphics[width=0.7\textwidth]{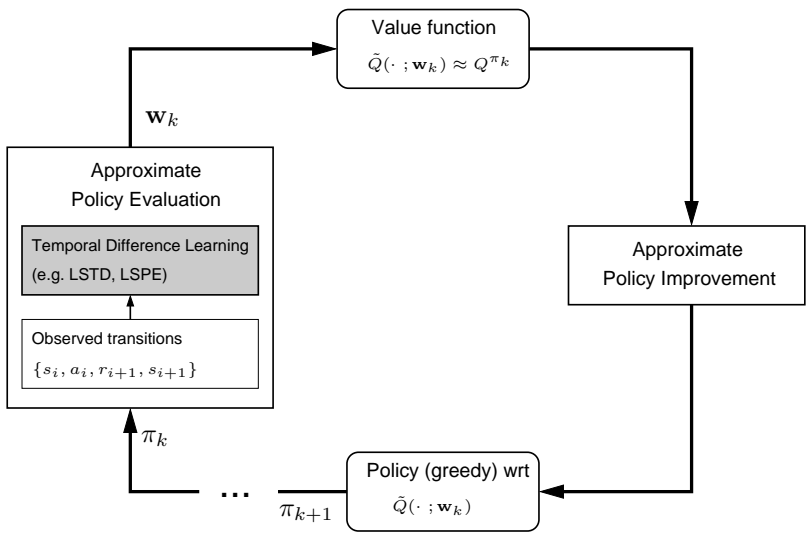}
	\caption{Approximate policy iteration framework.}
	\label{fig: API}
\end{figure}

However, in many practical situations this is not the case. If the number of states is very large or infinite, one
can only operate with an approximation of the Q-function, e.g. a linear approximation 
$\tilde Q(s,a;\bw)=\bphi_m(s,a)\trans\bw$, where $\bphi_m(s,a)$ is an $m$-dimensional feature vector 
and $\bw$ the adjustable weight vector. To approximate the unknown expectation value one employs
simulation (i.e.\ an agent interacts with the environment) to generate a large
number of observed transitions. Figure~\ref{fig: API} depicts the resulting approximate policy iteration 
framework: using only a parameterized $\tilde Q$ and sample transitions to emulate 
application of $\mathcal T_\pi$ means that we can carry out the policy evaluation step only approximately.
Also, using an approximation of $Q^{\pi_k}$ to derive an improved policy from does not necessarily mean that
the new policy actually is an improved one; oscillations in policy space are possible. In practice however, 
approximate policy iteration is a fairly sound procedure that either converges or oscillates
with bounded suboptimality \citep{bert96neurodynamicprogram}.

Inferring a parameter vector $\bw_k$ from sample transitions such that $\tilde Q(\cdot \ ;\bw_k)$ 
is a good approximation to $Q^{\pi_k}$ is therefore the central problem addressed by 
reinforcement learning. Chiefly two questions need to be answered: 
\begin{enumerate}
\item By what method do we choose the parametrisation of $\tilde Q$ and carry out regression?
\item By what method do we learn the weight vector $\bw$ of this approximation, given sample transitions?
\end{enumerate}
The latter can be solved by the family of temporal difference learning, with TD($\lambda$), initially 
proposed by \citet{sutton88td}, being its most prominent member.  Using a linearly parametrized
value function, it was in shown in \citep{tsivanroy97convergece_of_td} that TD($\lambda$) 
converges against the true value function (under certain technical assumptions).

\subsubsection{Approximate policy evaluation with least-squares methods}
In what follows we will discuss three related algorithms for approximate policy evaluation that share most of the advantages 
of TD($\lambda$) but converge much faster, 
since they are based on solving a least-squares problem in closed form, whereas TD($\lambda$) is based on
stochastic gradient descent. All three methods assume that an (infinitely) long\footnote{If we are dealing with
an episodic learning task with designated terminal states, we can generate an infinite trajectory
in the following way: once an episode ends, we set the discount factor $\gamma$ to zero
and make a zero-reward transition from the terminal state to the start state of the next (following)
episode.} trajectory of states and rewards is generated using a simulation of the system (e.g. an
agent interacting with its environment). The trajectory starts from an initial state $s_0$ and
consists of tuples $(s_0,a_0),(s_1,a_1),\ldots$ and rewards $r_1,r_2,\ldots$ where action $a_i$ is 
chosen according to $\pi$ and successor states and associated rewards are sampled from the underlying 
transition probabilities. From now on, to abbreviate these state-action tuples, we will understand $\bx_t$ as
denoting $\bx_t \definedas (s_t,a_t)$. Furthermore, we assume that the Q-function is parameterized by
 $\tilde Q^\pi(\bx;\bw)=\bphi_m(\bx)\trans\bw$ and that $\bw$ needs to be determined.

\paragraph{The LSPE($\lambda$) method.} 
The method $\lambda$-least squares policy evaluation LSPE($\lambda$) was 
proposed by \citet{nedicbert2003LSPE,BBN04@improved_td} and proceeds by making incremental changes to 
the weights $\bw$. Assume that at time $t$ (after having observed $t$ transitions) we have a current 
weight vector $\bw_t$ and observe a new transition from $\bx_t$ to $\bx_{t+1}$ with
associated reward $r_{t+1}$. Then we compute the solution $\hat \bw_{t+1}$ of the least-squares problem
\begin{equation}
\label{eq:LSPE1}
\bhatw_{t+1}=\argmin_{\bw} \sum_{i=0}^t\left\{ \bphi_m(\bx_i)\trans\bw - \bphi_m(\bx_i)\trans\bw_t - 
\sum_{k=i}^t (\lambda \gamma)^{k-i} d(\bx_k,\bx_{k+1};\bw_t) \right\}^2
\end{equation} 
where
\[
  d(\bx_k,\bx_{k+1};\bw_t) \definedas r_{k+1} + \gamma \bphi_m(\bx_{k+1})\trans\bw_t
      - \bphi_m(\bx_{k})\trans\bw_t.
\]
The new weight vector $\bw_{t+1}$ is obtained by setting 
\begin{equation}
\label{eq:LSPE1b}
\bw_{t+1}=\bw_t + \eta_t (\bhatw_{t+1} - \bw_t)
\end{equation}
where $\bw_0$ is the initial weight vector and $0<\eta_t\le 1$ is a diminishing step size.

\paragraph{The LSTD($\lambda$) method.} 
The least-squares temporal difference method LSTD($\lambda$) proposed by 
\citet{bradtke96lstd} for $\lambda=0$ and by \citet{boyan99lstd} for general $\lambda \in [0,1]$ does not proceed
by making incremental changes to the weight vector $\bw$. Instead, at time $t$ 
(after having observed $t$ transitions), the weight vector $\bw_{t+1}$
is obtained by solving the fixed-point equation
\begin{equation}
\label{eq:LSTD1}
\bhatw=\argmin_{\bw} \sum_{i=0}^t \left\{ \bphi_m(\bx_i)\trans\bw - \bphi_m(\bx_i)\trans\bhatw - 
\sum_{k=i}^t (\lambda \gamma)^{k-i} d(\bx_k,\bx_{k+1};\bhatw) \right\}^2
\end{equation} 
for $\bhatw$, where 
\[
  d(\bx_k,\bx_{k+1};\bhatw) \definedas r_{k+1} + \gamma \bphi_m(\bx_{k+1})\trans\bhatw
      - \bphi_m(\bx_{k})\trans\bhatw  ,
\]
and setting $\bw_{t+1}$ to this unique solution.

\paragraph{Comparison of LSPE and LSTD.}
The similarities and differences between LSPE($\lambda$) and LSTD($\lambda$) are listed in Table~\ref{tab:vergleich_von_lspe_lstd}. 
Both LSPE($\lambda$) and LSTD($\lambda$) converge to the same limit \citep[see][]{BBN04@improved_td}, which
is also the limit to which TD($\lambda$) converges (the initial iterates may be vastly different though).
Both methods rely on the solution of a least-squares problem (either explicitly 
as is the case in LSPE or implicitly as is the case in LSTD) and can be efficiently implemented using
recursive computations. Computational experiments in \citep{Ioffe96@lambdapi} or \citep{lagoudakis2003lspi} 
indicate that both approaches can perform much better than TD($\lambda$). 

Both methods LSPE and LSTD differ as far as their role in the approximate policy iteration framework is concerned. 
LSPE can take advantage of previous estimates of the weight vector and can hence be used in
the context of optimistic policy iteration (OPI), i.e.\ the policy under consideration gets improved
following very few observed transitions. For LSTD this is not possible; here a more rigid 
actor-critic approach is called for.

Both methods LSPE and LSTD also differ as far as their relation to standard regression with least-squares methods is 
concerned. LSPE directly minimizes a quadratic objective function. Using this function it will be possible to carry out `supervised' basis selection, where for the selection of basis functions the reduction of the costs (the quantity we are trying to minimize) is taken into account. For LSTD this is not possible; here
in fact we are solving a fixed point equation that employs least-squares only implicitly 
(to carry out the projection).    

\begin{table}
{\small 
\begin{center}
\begin{tabular}{|lll|}
\hline
\bfseries BRM& \bfseries LSTD & \bfseries LSPE\\
\hline
Corresponds to TD($0$) & Corresponds to TD($\lambda$) & Corresponds to TD($\lambda$)\\
Deterministic transitions only& Stochastic transitions possible & Stochastic transitions possible \\
No OPI & No OPI & OPI possible\\
Explicit least-squares & Least-squares only implicitly & Explicit least-squares\\
$\Rightarrow$ Supervised basis selection & $\Rightarrow$ No supervised basis selection & $\Rightarrow$ Supervised basis selection\\
\hline
\end{tabular}
\caption{Comparison of least-squares policy evaluation}
\label{tab:vergleich_von_lspe_lstd}
\end{center}
}
\end{table}

\paragraph{The BRM method.} 
A third approach, related to LSTD(0) is the
direct minimization of the Bellman residuals (BRM), as proposed in \citep{baird95residual,lagoudakis2003lspi}. 
Here, at time $t$, the weight vector $\bw_{t+1}$ is obtained from solving the least-squares problem 
\[ 
\bw_{t+1}=\argmin_{\bw} \sum_{i=0}^t \left\{ \bphi_m(\bx_i)\trans\bw - \sum_{s'} P(s'|s_i,\pi(s_i))
\left[ R(s'|s_i,\pi(s_i)) + \gamma \bphi_m(s',\pi(s'))\trans\bw \right]  \right\}^2
\]
Unfortunately, the transition probabilities can not be approximated by using single samples from
the trajectory; one would need `doubled' samples to obtain an unbiased estimate \citep[see][]{baird95residual}. Thus this method
would be only applicable for tasks with deterministic state transitions or known state dynamics; two  
conditions which are both violated
in our application to RoboCup-keepaway. Nevertheless 
we will treat the deterministic case in first place during all our derivations, since LSPE and 
LSTD require only very minor changes to the resulting implementation. Using BRM with 
deterministic transitions amounts to solving the least-squares problem
\begin{equation}
\label{eq:BRM}
\bw_{t+1}=\argmin_{\bw} \sum_{i=0}^t \left\{ \bphi_m(\bx_i)\trans\bw - r_{i+1} - \gamma \bphi_m(\bx_{i+1})\trans\bw   \right\}^2
\end{equation}

\subsection{Standard regression with regularization networks}
From the foregoing discussion we have seen that 
(approximate) policy evaluation can amount to a 
traditional function approximation problem. For this purpose we will here consider
the family of regularization networks \citep{RN95}, which are functionally
equivalent to kernel ridge regression and Bayesian regression with Gaussian
processes \citep{raswil06gpbook}. Here however, we will introduce them   
from the non-Bayesian regularization perspective as in \citep{smola2000SGMA}.

\subsubsection{Solving the full problem}
Given $t$ training examples $\{\bx_i,y_i\}_{i=1}^t$ with inputs $\bx_i$ and observed outputs $y_i$, 
to reconstruct the underlying function, 
one considers candidates from a function space $\mathcal H_k$, 
where $\mathcal H_k$ is a reproducing kernel Hilbert space  with reproducing
kernel $k$ \citep[e.g.][]{wahba1990}, and 
searches among all possible candidates for 
the function $f \in \mathcal H_k$ that achieves the minimum in the risk functional $\sum (y_i-f(\bx_i))^2
+ \sigma^2 \norm{f}_{\mathcal H_k}$. The scalar $\sigma^2$ is a regularization parameter.
Since solutions to this variational problem may be represented through the data alone \citep{wahba1990}
 as $f(\cdot)=\sum k(\bx_i,\cdot) w_i$, the unknown weight vector $\bw$ is obtained from solving the quadratic
problem 
\begin{equation}
\label{eq:full_problem}
\min_{\bw \in \mathbb R^t} \ (\bK \bw - \by)\trans(\bK \bw -\by) + \sigma^2 \bw\trans\bK \bw
\end{equation}  
The solution to \eqref{eq:full_problem} is $\bw=(\bK+\sigma^2 \bI)^{-1} \by$, where 
$\by=\rowvector{y_1,\ldots y_t}$ and $\bK$ is the $t \times t$ kernel matrix $[\bK]_{ij}=k(\bx_i,\bx_j)$.

\subsubsection{Subset of regressor approximation}
\label{sec:subset of regressors}
Often, one is not willing to solve the full $t$-by-$t$ problem in \eqref{eq:full_problem} when 
the number of training examples $t$ is large and instead considers means of approximation. In 
the subset of regressors (SR) approach \citep{poggiogirosi90rn,luowahba97has,smola2000SGMA} one chooses 
a subset
$\{\btx_i\}_{i=1}^m$ of the data, with $m \ll t$, and approximates the kernel 
for arbitrary $\bx,\bx'$ by taking
\begin{equation}
\label{eq:kernel_approx}
k(\bx,\bx')=\bkm{\bx}\trans \bKmmi  \bkm{\bx'}.
\end{equation}
Here $\bkm{\bx}$ denotes the $m \times 1$ feature vector 
$\bkm{\bx}=\rowvector{k(\btx_1,\bx),\ldots,k(\btx_m,\bx)}$
and the $m \times m$ matrix $\bKmm$ is the submatrix $[\bKmm]_{ij}=k(\btx_i,\btx_j)$ of 
the full kernel matrix $\bK$. Replacing the kernel in \eqref{eq:full_problem} by expression 
\eqref{eq:kernel_approx} gives 
\[
\min_{\bw \in \mathbb R^m} (\bKtm \bw - \by)\trans(\bKtm \bw -\by) + \sigma^2 \bw\trans\bKmm \bw
\]
with solution
\begin{equation}
\label{eq:SR_solution}
\bw_t=\bigl( \bKtm\trans \bKtm + \sigma^2 \bKmm\bigr)^{-1} \bKtm\trans \by
\end{equation}
where $\bKtm$ is the $t \times m$ submatrix $[\bKtm]_{ij}=k(\bx_i,\btx_j)$ corresponding 
to the $m$ columns of the data points in the subset. Learning 
the weight vector $\bw_t$ from \eqref{eq:SR_solution} costs $\mathcal O(tm^2)$ operations.
Afterwards, predictions for unknown test points $\bx_*$ are made by 
$f(\bx_*)=\bkm{\bx_*}\trans\bw$ at $\mathcal O(m)$ operations.

\subsubsection{Online selection of the subset}
\label{sect:SOG}
%
To choose the subset of relevant basis functions (termed the dictionary or set of basis vectors $\BV$)
many different approaches are possible; typically they can be distinguished as being unsupervised or supervised.
Unsupervised approaches like random selection \citep{williams01nystroem} or the incomplete Cholesky decomposition
\citep{fine01efficientsvmtrainin} do not use information about the task we want to solve, i.e.\ the response variable
we wish to regress upon. Random selection does not use any information at all whereas incomplete Cholesky
aims at reducing the error incurred from approximating the kernel matrix. Supervised choice of the subset
does take into account the response variable and usually proceeds by greedy forward selection, using e.g.
matching pursuit techniques \citep{smola01sparsegpr}.

However, none of these approaches are directly applicable for sequential learning, since the complete set
of basis function candidates must be known from the start. Instead, assume that the data becomes available
only sequentially at $t=1,2,\ldots$ and that only one pass over the data set is possible, so that we
cannot select the subset $\BV$ in advance. Working in the context of Gaussian process regression, 
\citet{csato2001sparse} and later
\citet{engel2003gptd} have proposed a sparse greedy online approximation: start from an empty set of $\BV$ and
examine at every time step $t$ if the new
example needs to be included in $\BV$ or if it can be processed without augmenting $\BV$. 
The criterion they employ to make that decision is an unsupervised one: at every time step $t$ compute
for the new data point $\bx_t$ the error
\begin{equation}
\label{eq:ALD-test}
\delta_t=k(\bx_t,\bx_t) - \bkm{\bx_t}\trans \bKmmi \bkm{\bx_t}
\end{equation}
incurred from approximating the new data point using the current $\BV$. If $\delta_t$ exceeds a given threshold
then it is considered as sufficiently different and added to the dictionary $\BV$. Note that only the current 
number of elements in $\BV$ at a given time $t$ is considered, the contribution from basis functions 
that will be added at a later time is ignored.

In this case, it might be instructive to visualize what happens to the $t \times m$ data 
matrix $\bKtm$ once $\BV$ is augmented. Adding the new element $\bx_t$ to $\BV$ means adding a new
basis function (centered on $\bx_t$) to the model and consequently adding a new associated column 
$\mathbf q=\rowvector{k(\bx_1,\bx_t),\ldots,k(\bx_t,\bx_t)}$ to $\bKtm$. With sparse online approximation all
$t-1$ past entries in $\mathbf q$ are given  
by $k(\bx_i,\bx_t)\approx \bkm{\bx_i}\trans \bKmmi \bkm{\bx_t}$, $i=1\ldots,t-1$, which is exact for the
$m$ basis-elements and an approximation for the remaining $t-m-1$ non-basis elements. Hence, going 
from $m$ to $m+1$ basis functions, we have that
\begin{equation}
\bK_{t,m+1}=\begin{bmatrix} \bK_{tm} & \mathbf q \end{bmatrix}=
\begin{bmatrix}
\bK_{t-1,m} & \bK_{t-1,m} \bat \\ \bkm{\bx_t}\trans & k(\bx_t,\bx_t)
\end{bmatrix}.
\end{equation}
where $\bat \definedas \bKmmi \bkm{\bx_t}$.
The overall effect is that now we do not need to access the full data set any longer. All costly 
$\mathcal O(tm)$ operations that arise from adding a new column, i.e.\ adding a new basis function,
computing the reduction of error during greedy forward selection of basis functions, or computing
predictive variance with augmentation as in \citep{rqc2005healing}, 
now become a more affordable $\mathcal O(m^2)$.

This is exploited in \citep{mein_icann2006}; here a simple modification of the
selection procedure is presented, where in addition to the unsupervised criterion 
from \eqref{eq:ALD-test} the contribution to the reduction of the error (i.e.\ the 
objective function one is trying to minimize) is taken into
account. Since the per-step complexity during training and then later during prediction critically 
depends on the size $m$ of the subset $\BV$, making sure that no unnecessary basis functions are 
selected ensures more efficient usage of otherwise scarce resources and makes learning in real-time
(a necessity for keepaway) possible.

\section{Policy evaluation with regularization networks}
\label{sec:pe with rn}
%
We now present an efficient online implementation for least-squares-based policy evaluation
(applicable to the methods LSPE, LSTD, BRM) to be used in the framework of approximate
policy iteration (see Figure~\ref{fig: API}). Our implementation 
combines the aforementioned automatic 
selection of basis functions (from Section~\ref{sect:SOG}) with a recursive computation of the weight vector
corresponding to the regularization network (from Section~\ref{sec:subset of regressors}) to represent the 
underlying Q-function. 
The goal is to infer an approximation $\tilde Q(\cdot \ ;\bw)$ of $Q^\pi$, the unknown Q-function 
of some given policy $\pi$. The training examples are taken from an observed trajectory
$\bx_0, \bx_1, \bx_2,\ldots$ with associated rewards $r_1,r_2,\ldots$ where $\bx_i$
denotes state-action tuples $\bx_i \definedas (s_i,a_i)$
and action $a_i=\pi(s_i)$ is selected according to policy $\pi$.

\subsection{Stating LSPE, LSTD and BRM with regularization networks}
First, express each of the three problems LSPE in eq.~\eqref{eq:LSPE1}, LSTD in eq.~\eqref{eq:LSTD1} and 
BRM in eq.~\eqref{eq:BRM} in more compact
matrix form using regularization networks from \eqref{eq:SR_solution}. Assume that the dictionary $\BV$ contains
$m$ basis functions. 
Further assume that at time $t$ (after having observed $t$ transitions) a new transition from $\bx_t$ to $\bx_{t+1}$
under reward $r_{t+1}$ is observed. From now on we will use a double index (also for vectors) to indicate the 
dependence in the number of examples $t$ and the number of basis functions $m$.
Define the matrices:
\begin{gather}
\bKttm\definedas 
\begin{bmatrix}  
    \bkm{\bx_0}\trans   \\
    \vdots \\
    \bkm{\bx_t}\trans 
    \end{bmatrix}, \quad 
\bHttm\definedas 
\begin{bmatrix}  
   \bkm{\bx_0}\trans-\gamma \bkm{\bx_1}\trans   \\
   \vdots \\
   \bkm{\bx_t}\trans-\gamma \bkm{\bx_{t+1}}\trans 
\end{bmatrix}  \bigskip  \nonumber \\
\brtt\definedas \begin{bmatrix} r_1 \\ \vdots\\ r_{t+1} \end{bmatrix}, \quad
\bLambda_{t+1}\definedas 
 \begin{bmatrix}
 1 & (\lambda \gamma)^1 & \cdots &(\lambda \gamma)^t \\
 0 &             \ddots     &        & \vdots \\
 \vdots  &  \ddots                & 1 & (\lambda \gamma)^1 \\
 0  & \cdots                  & 0  & 1
 \end{bmatrix}
 \label{eq:define_data_matrix}
\end{gather} 
where, as before, $m \times 1$ vector $\bkm{\cdot}$ denotes 
$\bkm{\cdot}=\rowvector{k(\cdot,\btx_1),\ldots,k(\cdot,\btx_m)}$.

\subsubsection{The LSPE($\lambda$) method}
Then, for LSPE($\lambda$), the least-squares problem \eqref{eq:LSPE1} is  stated as 
($\bwtm$ being the weight vector of the previous step):
\begin{eqnarray*}
\bhatw_{t+1,m}&=&\argmin_{\bw} \ \Bigl\{ \norm{\bKttm\bw - \bKttm \bwtm - \bLambda_{t+1}
\bigl(\brtt - \bHttm\bwtm\bigr)}^2 \nonumber \\
 & &  \qquad \qquad + \sigma^2 (\bw-\bwtm)\trans \bKmm (\bw-\bwtm) \Bigr\} 
\end{eqnarray*} 
Computing the derivative wrt $\bw$ and setting it to zero, one obtains for $\bhatw_{t+1,m}$:
\[
\bhatw_{t+1,m} =\bwtm + \bigl(\bKttm\trans \bKttm + \sigma^2 \bKmm\bigr)^{-1}
\bigl( \bZttm\trans \brtt - \bZttm\trans\bHttm\bwtm \bigr)
\]
where in the last line we have substituted $\bZttm\definedas \bLambda_{t+1}\trans \bKttm$.
From \eqref{eq:LSPE1b} the next iterate $\bwttm$ for the weight vector in LSPE($\lambda$) 
is thus obtained by
\begin{eqnarray}
\label{eq:LSPE3}
\bwttm&=&\bwtm + \eta_t (\bhatw_{t+1,m} - \bwtm)=\bwtm+\eta_t 
\bigl(\bKttm\trans \bKttm + \sigma^2 \bKmm\bigr)^{-1} \nonumber \\
& & 
\bigl( \bZttm\trans \brtt - \bZttm\trans\bHttm \bwtm \bigr)
\end{eqnarray}

\subsubsection{The LSTD($\lambda$) method}
Likewise, for LSTD($\lambda$), the fixed point equation \eqref{eq:LSTD1} is stated as:
\begin{eqnarray*}
\bhatw &=&\argmin_{\bw} \ \Bigl\{ \norm{\bKttm\bw - \bKttm \bhatw - \bLambda_{t+1}
\bigl(\brtt - \bHttm\bhatw\bigr)}^2 \nonumber \\
 & & \qquad \qquad + \sigma^2 \bw\trans \bKmm \bw \Bigr\}.
\end{eqnarray*}
Computing the derivative with respect to $\bw$ and setting it to zero, one obtains 
\[ \bigl( \bZttm\trans \bHttm + \sigma^2 \bKmm \bigr) \bhatw=\bZttm\trans \brtt. \]
Thus the solution $\bwttm$ to the fixed point equation in LSTD($\lambda$) is 
given by:
\begin{equation}
\label{eq:LSTD3}
\bwttm=\bigl( \bZttm\trans\bHttm + \sigma^2 \bKmm \bigr)^{-1} \bZttm\trans \brtt
\end{equation}

\subsubsection{The BRM method}
Finally, for the case of BRM, the least-squares problem \eqref{eq:BRM} is stated as:
\[
\bwttm=\argmin_\bw \ \Bigl\{ \norm{\brtt- \bHttm\bw }^2 + \sigma^2 \bw\trans \bKmm \bw \Bigr\} 
\]
Thus again, one obtains the weight vector $\bwttm$ by
\begin{equation}
\label{eq:BRM2}
\bwttm=\bigl( \bHttm\trans\bHttm + \sigma^2 \bKmm \bigr)^{-1} \bHttm\trans \brtt
\end{equation}

\subsection{Outline of the recursive implementation}
Note that all three methods amount to solving a very similar structured set of linear 
equations in eqs. \eqref{eq:LSPE3},\eqref{eq:LSTD3},\eqref{eq:BRM2}. Overloading the notation
these can be compactly stated as:
\begin{itemize}
	\item {\bfseries LSPE:} solve 
	  \begin{equation}
	       \bwttm=\bwtm + \eta \bPittm(\bbttm-\bAttm\bwtm) \tag{\ref{eq:LSPE3}'} 
	  \end{equation}
	  where 
	  \begin{itemize}
	     \item $\bPittm \definedas (\bKttm\trans\bKttm + \sigma^2 \bKmm)^{-1}$
	     \item $\bbttm \definedas \bZttm\trans\brtt$
	     \item $\bAttm \definedas \bZttm\trans\bHttm$
	  \end{itemize}
	\item {\bfseries LSTD:} solve 
	    \begin{equation}
	         \bwttm=\bPittm\bbttm \tag{\ref{eq:LSTD3}'}
	     \end{equation}
	     where 
	     \begin{itemize}
	       \item $\bPittm \definedas (\bZttm\trans\bHttm + \sigma^2 \bKmm)^{-1}$
	       \item $\bbttm \definedas \bZttm\trans\brtt$
	     \end{itemize}
	\item {\bfseries BRM:} solve 
	     \begin{equation}
	       \bwttm=\bPittm\bbttm \tag{\ref{eq:BRM2}'}
	     \end{equation}
	     where 
	     \begin{itemize}
	       \item $\bPittm \definedas (\bHttm\trans\bHttm + \sigma^2 \bKmm)^{-1}$
	       \item $\bbttm \definedas \bHttm\trans\brtt$
	     \end{itemize}	 	     
\end{itemize}
Each time a new transitions from $\bxt$ to $\bxtt$ under reward $\rtt$ is observed, the goal is to recursively 
\begin{enumerate}
  \item update the weight vector $\bwtm$, and
  \item possibly augment the model, adding a new basis function (centered on $\bxtt$) to the set of 
  currently selected basis functions $\BV$. 
\end{enumerate}
More
specifically, we will perform one or both of the following update operations:
\begin{enumerate}
\item {\em Normal step}: Process $(\bxtt,\rtt)$ using the current fixed set of 
      basis functions $\BV$.
\item {\em Growing step}: If the new example is sufficiently different from the previous examples in $\BV$
      (i.e.\ the reconstruction error in (\ref{eq:ALD-test}) exceeds a given threshold)
      and strongly contributes to the solution of the problem (i.e.\ the decrease of the
      loss when adding the new basis function is greater than a given threshold) 
      then the current example is added to $\BV$ and the number of basis functions
      in the model is increased by one.
\end{enumerate} 
The update operations work along the lines of recursive least squares (RLS), i.e.\ 
propagate forward the inverse\footnote{A better alternative (from the standpoint of
numerical implementation) would be to not propagate forward the inverse, but instead to work with 
the Cholesky factor. For this paper we chose the first method in the first place because it gives consistent
update formulas for all three considered problems (note that for LSTD the cross-product matrix is not symmetric)  and overall allows a better exposition. For details on the second way, see e.g. \citep{sayed03adaptivefiltering}.} of the $m \times m$ cross product matrix $\bPtm$. 
Integral to the derivation of these updates are two well-known matrix identities for recursively computing the 
inverse of a matrix: (for matrices with compatible dimensions)    
\begin{equation}
 \label{eq:SMW}
 \text{if } \bB_{t+1}=\bB_t+\bb\bb\trans
 \text{ then }
  \bB_{t+1}^{-1}=\bB_t^{-1} - \frac{\bB_t^{-1}\bb\bb\trans\bB_t^{-1}}{1+\bb\trans \bB_t^{-1} \bb}
\end{equation}
which is used when adding a row to the data matrix. Likewise,
\begin{equation}
\label{eq:PMI}
\text{if } \bB_{t+1}= \begin{bmatrix} \bB_t & \bb \\ \bb\trans & b^* \end{bmatrix}
\text{ then }
 \bB_{t+1}^{-1}= 
   \begin{bmatrix}\bB_t^{-1} & \bzeros \\ \bzeros & 0 \end{bmatrix} + \frac{1}{\Delta_b}
   \begin{bmatrix}-\bB_t^{-1}\bb \\ 1\end{bmatrix} 
   \begin{bmatrix}-\bB_t^{-1}\bb \\ 1\end{bmatrix}\trans
\end{equation}
with $\Delta_b=\bstar-\bb\trans \bB_t^{-1}\bb$. This second update is used when adding a column
to the data matrix.

\begin{figure*}[tbh]
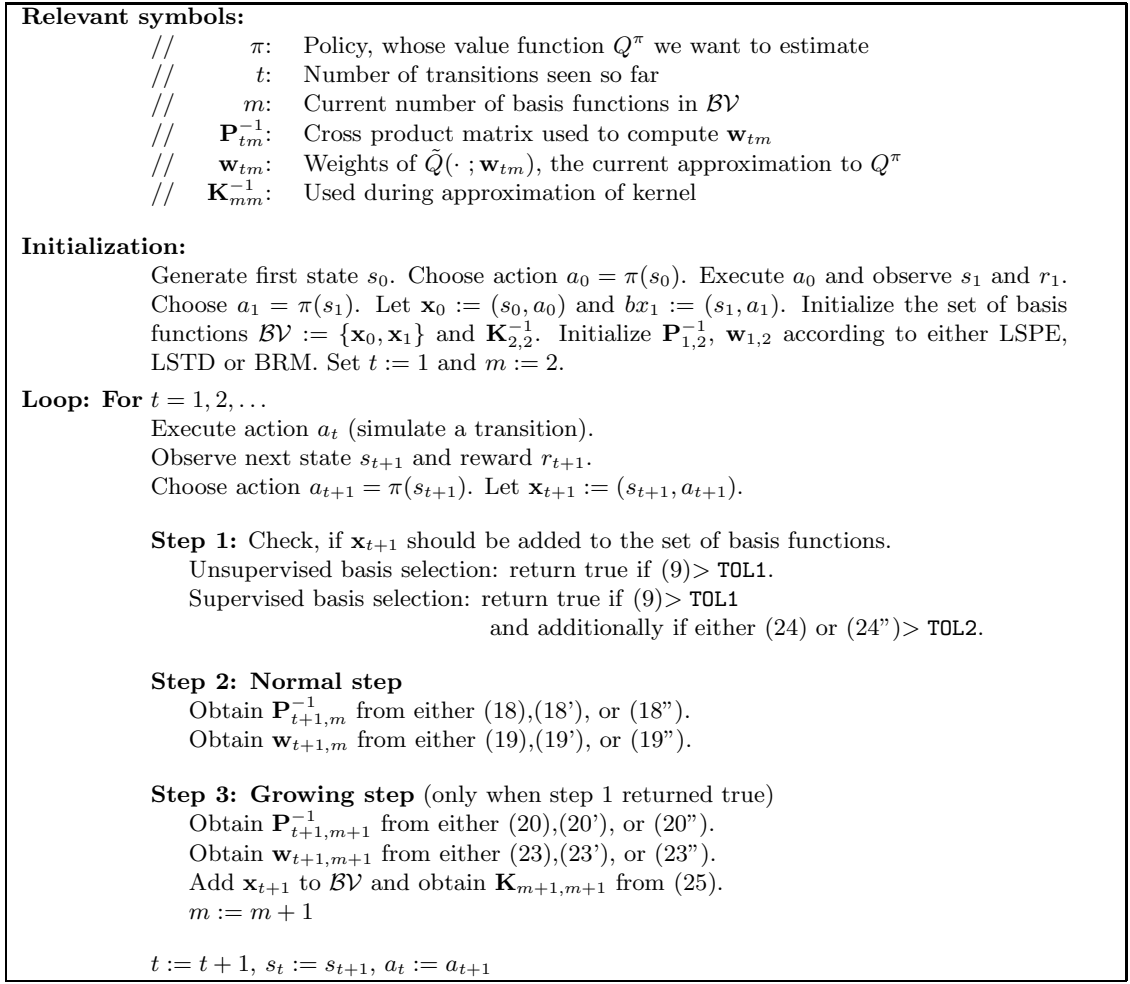

{\footnotesize
\begin{center}
\begin{tabular}{|p{0.95\textwidth}|}
\hline
{\bf Relevant symbols:} \\
\hspace*{1.4cm} \begin{tabular}{rrl}
//& $\pi$: & Policy, whose value function $Q^\pi$ we want to estimate\\  
//& $t$: & Number of transitions seen so far \\
//& $m$: & Current number of basis functions in $\BV$ \\
//& $\bPitm$: & Cross product matrix used to compute $\bwtm$ \\
//& $\bwtm$: & Weights of $\tilde Q(\cdot \ ;\bwtm)$, the current approximation to $Q^\pi$\\
//& $\bKmmi$: & Used during approximation of kernel\\
\end{tabular} \\
\smallskip
{\bf Initialization:}\\
\hspace*{1.4cm}  \begin{tabular}{p{0.8\textwidth}}
                       Generate first state $s_0$. Choose action $a_0=\pi(s_0)$. Execute $a_0$ and observe
                       $s_1$ and $r_1$. Choose $a_1=\pi(s_1)$. Let $\bx_0\definedas(s_0,a_0)$ and 
                       $bx_1\definedas(s_1,a_1)$. Initialize the set of basis functions 
                       $\BV\definedas\{\bx_0,\bx_1\}$ and $\bK_{2,2}^{-1}$. Initialize $\bP_{1,2}^{-1}$,
                       $\bw_{1,2}$ according to either LSPE, LSTD or BRM. Set $t\definedas 1$ and
                       $m \definedas 2$. 
                  \end{tabular} \smallskip \\ 
{\bf Loop:} {\bfseries For} $t=1,2,\ldots$  \\
\hspace*{1.4cm}  \begin{tabular}{p{0.8\textwidth}}
                    Execute action $a_t$ (simulate a transition). \\
                    Observe next state $s_{t+1}$ and reward $r_{t+1}$. \\
                    Choose action $a_{t+1}=\pi(s_{t+1})$. Let $\bx_{t+1}\definedas(s_{t+1},a_{t+1})$. \\ \smallskip
                    {\bfseries Step 1:} Check, if $\bx_{t+1}$ should be added to the set of basis functions. \\
                     \hspace*{0.4cm} Unsupervised basis selection: return true if \eqref{eq:ALD-test}$>\texttt{TOL1}$.\\
                     \hspace*{0.4cm} Supervised basis selection: return true if \eqref{eq:ALD-test}$>\texttt{TOL1}$\\
                     \hspace*{4.4cm} and additionally if either \eqref{eq:xitmm} or (\ref{eq:xitmm}'')$>\texttt{TOL2}$. \\ \smallskip
                    {\bfseries Step 2: Normal step} \\
                     \hspace*{0.4cm} Obtain $\bPittm$ from either
                                      \eqref{eq:normal Pittm},(\ref{eq:normal Pittm}'), or (\ref{eq:normal Pittm}''). \\ 
                     \hspace*{0.4cm} Obtain $\bwttm$ from either 
                                      \eqref{eq:normal wttm},(\ref{eq:normal wttm}'), or (\ref{eq:normal wttm}''). \\ \smallskip
                    {\bfseries Step 3: Growing step} (only when step 1 returned true) \\
                       \hspace*{0.4cm} Obtain $\bPittmm$ from either 
                                      \eqref{eq:Pitmm},(\ref{eq:Pitmm}'), or (\ref{eq:Pitmm}''). \\
                       \hspace*{0.4cm} Obtain $\bwttmm$ from either
                                       \eqref{eq:bbetatmm},(\ref{eq:bbetatmm}'), or (\ref{eq:bbetatmm}''). \\                
                       \hspace*{0.4cm} Add $\bxtt$ to $\BV$ and obtain $\bK_{m+1,m+1}$ from  
                       								\eqref{eq: aktualisiere Kmm}. \\
                       \hspace*{0.4cm} $m:=m+1$ \\ \smallskip
                       $t:=t+1$, $s_t:=s_{t+1}$, $a_t:=a_{t+1}$\\						                       
                 \end{tabular} \\                
\hline
\end{tabular}
\end{center}
}
\caption{Online policy evalution with growing regularization networks. This pseudo-code applies to BRM, LSPE 
and LSTD, see the appendix for the exact equations. The computational complexity per observed 
transition is $\mathcal O(m^2)$.}
\label{fig:algorithm}
\end{figure*}

An outline of the general implementation applicable to all three of the methods LSPE, LSTD, and BRM is
sketched in Figure~\ref{fig:algorithm}. To avoid unnecessary repetitions we will here only derive the update equations
for the BRM method; the other two are obtained with very minor modifications and are summarized in
the appendix.

\subsection{Derivation of recursive updates for the case BRM}
Let $t$ be the current time step, $(\bxtt,\rtt)$ the currently observed input-output pair and assume 
that from the past $t$ examples $\{(\bx_i,r_i)\}_{i=1}^t$ the $m$ 
examples $\{\tilde \bx_i\}_{i=1}^m$ were selected into the dictionary $\BV$. Consider the penalized
least-squares problem that is BRM (restated here for clarity) 
\begin{equation}
\label{eq:BRM4}
  \min_{\bw \in \mathbb R^m} \Jtm(\bw) = \norm{\brt - \bHtm\bw}^2 + \sigma^2 \bw\trans \bKmm \bw
\end{equation}
with $\bHtm$ being the $t \times m$ data matrix  and $\brt$ being the $t \times 1$ vector of the observed output values from \eqref{eq:define_data_matrix}. Defining the $m \times m$ cross product matrix 
$\bPtm=(\bHtm\trans \bHtm + \sigma^2 \bKmm)$, the solution to (\ref{eq:BRM4}) is given 
by 
\[ 
   \bwtm=\bPitm \bHtm\trans \brt.
\] 
Finally, introduce the costs $\xitm=\Jtm(\bwtm)$. Assuming that 
$\{\bPitm,\ \bwtm,\ \xitm\}$ are known from previous computations, every time a new transition
$(\bxtt,\rtt)$ is observed, we will perform one or both of the following update operations:
%
%

\subsubsection{Normal step: from $\{\bPitm,\bwtm,\xitm\}$ to $\{\bPittm,\bwttm,\xittm\}$}
\label{sect:normalstep}
With $\bhtt$  defined as $\bhtt\definedas \rowvector{\bkm{\bxt}-\gamma \bkm{\bxtt}}$, one gets
\[
\bHttm=\begin{bmatrix} \bHtm \\ \bhtt\trans \end{bmatrix} \quad \text{and} \quad 
\brtt=\begin{bmatrix} \brt \\ \rtt \end{bmatrix}.
\]
Thus $\bPttm=\bPtm+\bhtt\bhtt\trans$ and we obtain from (\ref{eq:SMW}) the 
well-known RLS updates
\begin{equation}
\label{eq:normal Pittm}
\bPittm  =  \bPitm - \frac{\bPitm\bhtt \bhtt\trans\bPitm}{\Delta} 
\end{equation}
with scalar $\Delta=1+\bhtt\trans \bPitm \bhtt$ and 
\begin{equation}
\label{eq:normal wttm}
    \bwttm  =  \bwtm + \frac{\varrho}{\Delta} \bPitm \bhtt 
\end{equation}
with scalar $\varrho=\rtt - \bhtt\trans \bwtm$. The costs become 
$\xittm  = \xitm + \frac{\varrho^2}{\Delta}$. The set of basis 
functions $\BV$ is not altered during this step. Operation complexity is $\mathcal O(m^2)$.

\subsubsection{Growing step: from $\{\bPittm,\bwttm,\xittm\}$ to $\{\bPittmm,\bwttmm,\xittmm\}$}
\label{sect:growing step}

\paragraph{How to add a $\BV$.}
When adding a basis function (centered on $\bxtt$) to 
the model, 
we augment the set $\BV$ with $\btx_{m+1}$ (note that $\btx_{m+1}$ is the same as $\bxtt$ from above).
Define $\bktt\definedas \bkm{\btx_{m+1}}$, $\kstar_t \definedas k(\bxt,\bxtt)$, 
and $\kstar_{t+1}\definedas k(\bxtt,\bxtt)$.
Adding a basis function means appending a new $(t+1) \times 1$ vector $\bq$ to the data matrix and appending 
$\bktt$ as row/column to the penalty matrix $\bKmm$, thus
\[
 \bPttmm=\begin{bmatrix} \bHttm & \bq \end{bmatrix}\trans \begin{bmatrix} \bHttm & \bq \end{bmatrix}
        + \sigma^2 \begin{bmatrix} \bKmm & \bktt \\ \bktt\trans & \kstar_{t+1} \end{bmatrix}.
\] 
Invoking (\ref{eq:PMI}) we obtain the updated inverse $\bPittmm$ via
\begin{equation}
\label{eq:Pitmm}
\bPittmm=\begin{bmatrix} \bPittm & \bzeros \\ \bzeros & 0 \end{bmatrix} + 
    \frac{1}{\Delta_b} \begin{bmatrix} -\bwb \\ 1 \end{bmatrix}
    \begin{bmatrix} -\bwb \\ 1 \end{bmatrix}\trans
\end{equation}
where simple vector algebra reveals that
\begin{align}
\label{eq:wbdeltab}
\bwb&=\bPittm (\bHttm\trans \bq + \sigma^2 \bktt) \nonumber \\
\Delta_b&=\bq\trans \bq + \sigma^2 \kstar_{t+1} - (\bHttm\trans \bq + \sigma^2 \bktt)\trans \bwb.
\end{align}
Without sparse online approximation this step 
would require us to recall all $t$ past examples and would come at the 
undesirable price of $\mathcal O(tm)$ operations.
However, we are going to get away with merely $\mathcal O(m)$ operations and 
only need to access the $m$ past examples in the memorized $\BV$. 
Due to the sparse online approximation, $\bq$ is actually of the form 
$ \bq\trans= \begin{bmatrix} \bHtm \batt & \ \  \hstar_{t+1} \end{bmatrix}\trans$
with $\hstar_{t+1} \definedas \kstar_t-\gamma \kstar_{t+1}$ and $\batt=\bKmm^{-1} \bktt$ 
(see Section~\ref{sect:SOG}). Hence new information is injected 
only through the last component. Exploiting this special structure of $\bq$ equation
(\ref{eq:wbdeltab}) becomes 
\begin{align}
\label{eq:wbdeltab2}
\bwb& =\batt + \frac{\delta_h}{\Delta} \bPitm \bhtt \nonumber \\
\Delta_b& = \frac{\delta_h^2}{\Delta}+\sigma^2\delta_h
\end{align}
where $\delta_h=\hstar_{t+1} - \bhtt\trans \batt$. If we cache
and reuse those terms already computed in the preceding step 
(see Section~\ref{sect:normalstep}) then we can obtain $\bwb, \Delta_b$ in
$\mathcal O(m)$ operations.

To obtain the updated coefficients $\bwttmm$ we postmultiply (\ref{eq:Pitmm})
by 
$\bHttmm\trans \brtt=\begin{bmatrix} \bHttm\trans \brtt & \  \bq\trans\brtt\end{bmatrix}\trans$, 
getting 
\begin{equation}
\label{eq:bbetatmm}
\bwttmm=\begin{bmatrix} \bwtm \\ 0 \end{bmatrix} + \kappa 
 \begin{bmatrix} -\bwb \\ 1 \end{bmatrix}
\end{equation} 
where scalar $\kappa$ is defined by $\kappa=\brtt\trans(\bq-\bHttm\bwb) / \Delta_b$.
Again we can now exploit the special structure of $\bq$ to show that $\kappa$
is equal to 
\[
\kappa=-\frac{\delta_h\varrho}{\Delta_b\Delta}
\]
And again we can reuse terms computed in the previous step (see Section~ \ref{sect:normalstep}).

Skipping the  computations, we can show that the reduced (regularized)
cost $\xittmm$ is recursively obtained from $\xittm$ via the expression:
\begin{equation}
\label{eq:xitmm}
\xittmm=\xittm - \kappa^2 \Delta_b.
\end{equation}
Finally, each time we add an example to the $\BV$ set we must also update the
inverse kernel matrix $\bKmm^{-1}$ needed during the computation of $\batt$ and
$\delta_h$. This can be done using the formula for partitioned matrix inverses 
(\ref{eq:PMI}):
\begin{equation}
\label{eq: aktualisiere Kmm}
\bK_{m+1,m+1}^{-1}=\begin{bmatrix} \bKmmi & \bzeros \\ \bzeros\trans & 0\end{bmatrix}
+\frac{1}{\delta}\begin{bmatrix} -\batt \\ 1 \end{bmatrix}
\begin{bmatrix} -\batt \\ 1 \end{bmatrix}\trans.
\end{equation}

\medskip

\paragraph{When to add a $\BV$.}
To decide whether or not the current example $\bxtt$ should be added to the $\BV$ set,
we employ the supervised two-part criterion from \citep{mein_icann2006}. 
The first part measures the `novelty' of the current 
example: only examples that are `far' from those already stored in the $\BV$ set
are considered for inclusion. To this end we compute as in 
\citep{csato2001sparse} the squared norm of the residual from
projecting (in RKHS) the example onto the span of the current $\BV$ set, i.e.\ 
we compute, restated from (\ref{eq:ALD-test}), 
$\delta=\kstar_{t+1}-\bktt\trans\batt$. 
If $\delta<\mathtt{TOL1}$ for a given threshold $\mathtt{TOL1}$, 
then $\bxtt$ is well represented by the given $\BV$ set
and its inclusion would not contribute much to reduce the error from approximating
the kernel by the reduced set. On the other hand, if $\delta>\mathtt{TOL1}$ then
$\bxtt$ is not well represented by the current $\BV$ set and leaving it
behind could incur a large error in the approximation of the kernel.

Aside from novelty, we consider as second part 
of the selection criterion the `usefulness' of a basis function candidate.  
Usefulness is taken to be its contribution to the reduction of the regularized
costs $\xitm$, i.e.\ the term $\kappa^2\Delta_b$ from (\ref{eq:xitmm}). Both parts together
are combined into one rule: only if $\delta > \mathtt{TOL1}$ and 
$ \delta \kappa^2 \Delta_b > \mathtt{TOL2}$,
then the current example will become a new basis function and will be added to $\BV$.

\section{RoboCup-keepaway as RL benchmark}
\label{sec:robocup}
The experimental work we carried out for this article uses the publicly available\footnote{Sources are available 
from {\tt http://www.cs.utexas.edu/users/AustinVilla/sim/keepaway/}.} keepaway 
framework from \citep{AB05}, which is built on top of the standard RoboCup soccer simulator
also used for official competitions \citep{noda98ss}. Agents in RoboCup are autonomous entities; they 
sense and act independently and asynchronously, run as individual processes and cannot communicate directly. 
Agents receive visual perceptions every 150 msec and may act once every 100 msec. 
The state description consists of relative distances and angles 
to visible objects in the world, such as the ball, other agents or fixed beacons
for localization. In addition, random noise affects both the agents sensors as well as their actuators.

In keepaway, one team of `keepers' must learn how to maximize the time they can control the ball within a 
limited region of the field against an opposing team of `takers'. Only the keepers are allowed to learn, 
the behavior of the takers is governed 
by a fixed set of hand-coded rules. However, each keeper only learns {\em individually} 
from its own (noisy) actions
and its own (noisy) perceptions of the world. The decision-making happens at an intermediate level
using multi-step macro-actions; the keeper currently controlling the ball must decide between holding
the ball or passing it to one of its teammates. The remaining keepers automatically try to position
themselves such to best receive a pass. The task is episodic; it starts 
with the keepers controlling the ball and continues as long as neither the ball leaves the
region nor the takers succeed in gaining control. Thus the goal for RL is to maximize the overall duration
of an episode. The immediate reward is the time that passes between individual calls
to the acting agent.

\begin{figure}[tb]
	\centering
		\includegraphics[height=4.5cm]{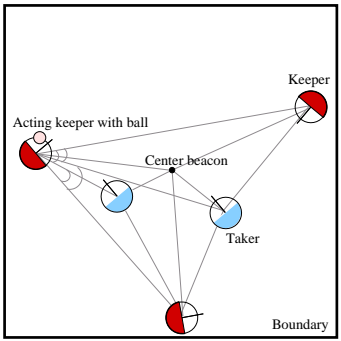}
	\caption{Illustrating {\em keepaway}. The various lines and angles indicate the 13 state variables 
	 making up each sensation as provided by the keepaway benchmark software.}
	\label{fig:keepaway}
\end{figure}

For our work, we consider as in \citep{AB05} the special 3vs2 
keepaway problem (i.e.\ three learning keepers against two takers) played in a 20x20m field.
In this case the continuous state space has dimensionality 13, and the discrete action space consists of the 
three different actions {\em hold, pass to teammate-1, pass to teammate-2} (see Figure~\ref{fig:keepaway}). 
More generally, larger instantiations of keepaway would also be possible, like e.g. 4vs3, 5vs4 or more, 
resulting in even larger state- and action spaces.     

\section{Experiments}
\label{sec:experiments and results}
In this section we are finally ready to apply our proposed approach to the keepaway problem. We implemented 
and compared two different variations of the basic algorithm in a policy iteration based framework: 
(a) Optimistic policy iteration using LSPE($\lambda$) and (b) Actor-critic policy iteration using LSTD($\lambda$).
As baseline method we used Sarsa($\lambda$) with tilecoding, which we re-implemented from \citep{AB05}
as faithfully as possible. Initially, we also tried to employ BRM instead of LSTD in the actor-critic framework. However,
this set-up did not fare well in our experiments because of the stochastic state-transitions in keepaway
(resulting in highly variable outcomes) and BRM's inability to deal with this situation adequately. Thus, the results
for BRM are not reported here.

\paragraph{Optimistic policy iteration.} 
Sarsa($\lambda$) and LSPE($\lambda$) paired with optimistic policy iteration is an on-policy learning method, meaning that the 
learning procedure estimates the Q-values from and for the current policy being executed by the agent. At the same time, 
the agent continually updates the policy according to the changing estimates of the Q-function. Thus 
policy evaluation and improvement are tightly interwoven. Optimistic policy iteration (OPI) is an online method
that immediately processes the observed transitions as they become available from the agent interacting with 
the environment \citep{bert96neurodynamicprogram,sutton98introduction}.

\paragraph{Actor-critic.}
In contrast, LSTD($\lambda$)
paired with actor-critic is an off-policy learning method adhering with more rigor to the policy iteration framework. 
Here the learning procedure estimates the Q-values
for a fixed policy, i.e.\ a policy that is not continually modified to reflect the changing estimates of Q.
Instead, one collects a large number of state transitions under the same policy and estimates Q from these 
training examples. In OPI, where the most recent version of the Q-function is used to derive the next control action, 
only one network is required to represent Q and make the predictions. In contrast, the actor-critic framework maintains 
two instantiations of regularization networks: one (the actor) is used to represent the Q-function learned during the  
previous policy evaluation step and which is now used to represent the current policy, i.e.\ control actions are
derived using its predictions. The second network (the critic) is used to represent the current Q-function and
is updated regularly. 

One advantage of the actor-critic approach is that we can reuse the same set of
observed transitions to evaluate different policies, as proposed in \citep{lagoudakis2003lspi}.
 We maintain an ever-growing list of all transitions observed from the learning agent (irrespective
of the policy), and use it to evaluate the current policy with LSTD($\lambda$). To reflect the real-time nature of learning in RoboCup, where 
we can only carry out a very small amount of computations during one single function call to the agent, we evaluate the
transitions in small batches (20 examples per step). Once we have completed evaluating all training examples in the list, the
critic network is copied to the actor network and we can proceed to the next iteration, starting anew to process the examples,
using this time a new policy.  
    
\paragraph{Policy improvement and $\varepsilon$-greedy action selection.}
To carry out policy improvement, every time we need to determine a control action for an arbitrary state
$s^*$, we choose the action $a^*$ that achieves the maximum Q-value; that is, given weights $\bw_k$ 
and a set of basis functions $\{\btx_1,\ldots,\btx_m\}$, we choose
\[
  a^*=\argmax_{a} \ \tilde Q(s^*,a;\bw_k)=\argmax_{a} \ \bkm{s^*,a}\trans\bw_k.
\]
Sometimes however, instead of choosing the best (greedy) action, it is recommended to try out
an alternative (non-greedy) action to ensure sufficient exploration. Here we employ the 
$\varepsilon$-greedy selection scheme; we choose a random action with a small probability 
$\varepsilon$ ( $\varepsilon=0.01$), otherwise we pick the greedy action with probability 
$1-\varepsilon$. Taking a random action usually means to choose among all possible actions
with equal probability.

Under the standard assumption for errors in Bayesian regression 
\citep[e.g., see][]{raswil06gpbook}, namely that the observed target values differ from
the true function values by an additive noise term (i.i.d.
Gaussian noise with zero mean and uniform variance), it is also possible to obtain an 
expression for the `predictive variance' which measures the uncertainty associated
with value predictions. The availability of such confidence intervals (which is possible
for the direct least-squares problems LSPE and also BRM) could be used, as suggested
in \citep{engel2005rlgptd}, to guide the choice of actions during exploration and to increase
the overall performance.  For the purpose of solving the keepaway problem 
however, our initial experiments showed no measurable increase in performance
when including this additional feature.

\paragraph{Remaining parameters.}
Since the kernel is defined for state-action tuples, we employ a product kernel $k([s,a],[s',a'])=k_S(s,s')k_A(a,a')$ as suggested
by \citet{engel2005rlgptd}. The action kernel $k_A(a,a')$ is taken to be the Kronecker delta, since the actions in keepaway
are discrete and disparate. As state kernel $k_S(s,s')$ we chose the Gaussian RBF $k_S(s,s')=\exp(-h\norm{s-s'}^2)$ 
with uniform length-scale $h^{-1}=0.2$. The other
parameters were set to: regularization $\sigma^2=0.1$, discount factor for 
RL $\gamma=0.99$, $\lambda=0.5$, and LSPE step size $\eta_t=0.5$. The novelty parameter for basis selection was set to
 $\texttt{TOL1}=0.1$.
For the usefulness part we  tried out different values to examine the effect supervised basis selection has;  
we started with $\texttt{TOL2}=0$ corresponding to the unsupervised case and then began increasing the tolerance, considering
 alternatively the settings $\texttt{TOL2}=0.001$  and $\texttt{TOL2}=0.01$.
Since in the case of LSTD we are not directly solving a least-squares problem, we use the associated BRM formulation to
obtain an expression for the error reduction in the supervised basis selection. 
Due to the very long runtime of the simulations (simulating one hour in the soccer server roughly takes one hour
real time on a standard PC) we could not try out many different parameter combinations.       
The parameters governing RL were set according to our experiences with smaller problems and are in the range 
typically reported in the literature.
The parameters governing the choice of the  kernel (i.e.\ the length-scale of the Gaussian RBF) was chosen such 
that for the unsupervised case ($\texttt{TOL2}=0$) the number of selected basis functions approaches 
the maximum number of basis functions the CPU used for these the experiments was able to process in real-time. This number
was determined to be $\sim 1400$ (on a standard 2 GHz PC).

\begin{figure}[hp]
	\centering
\includegraphics[width=0.47\textwidth]{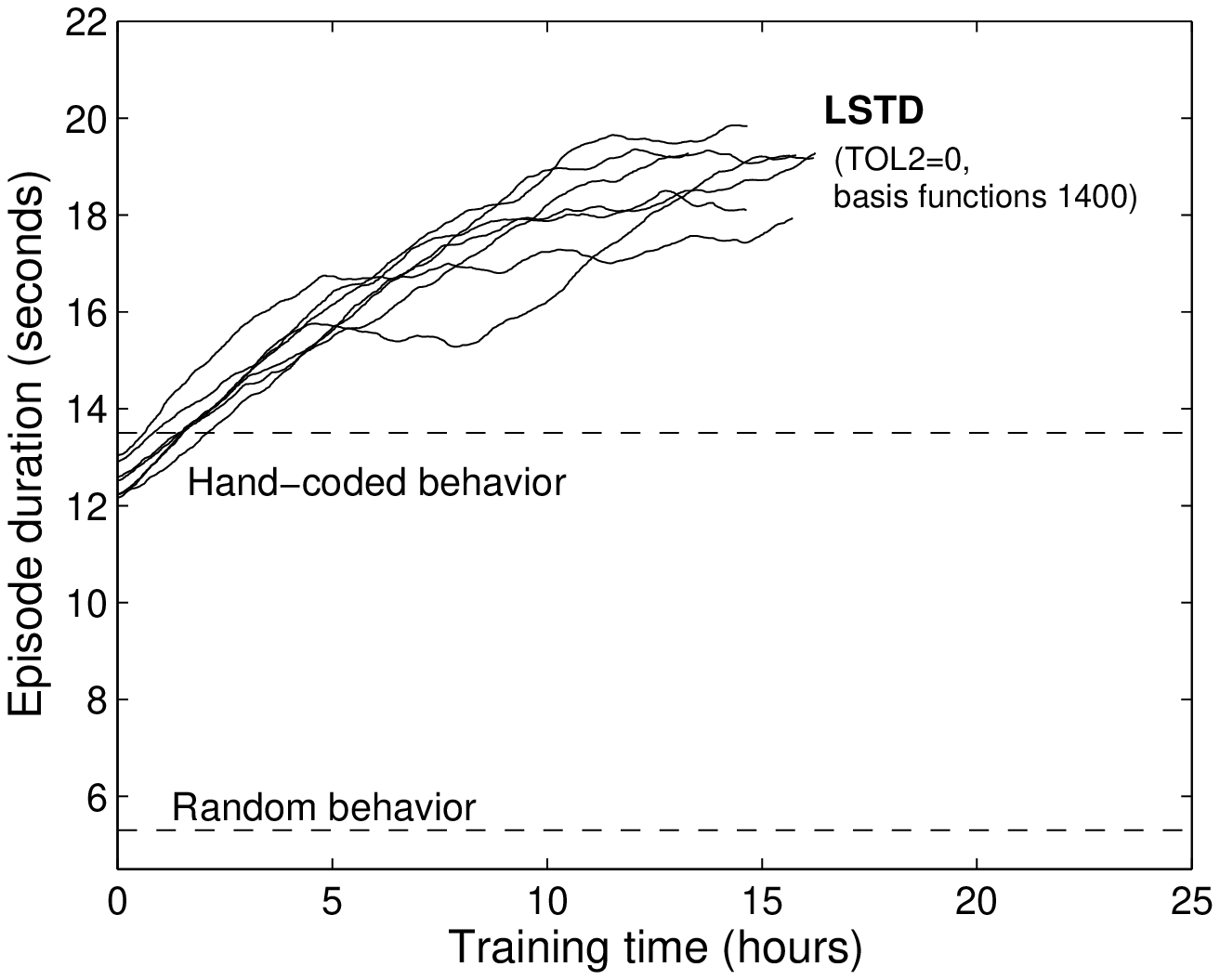}
\includegraphics[width=0.47\textwidth]{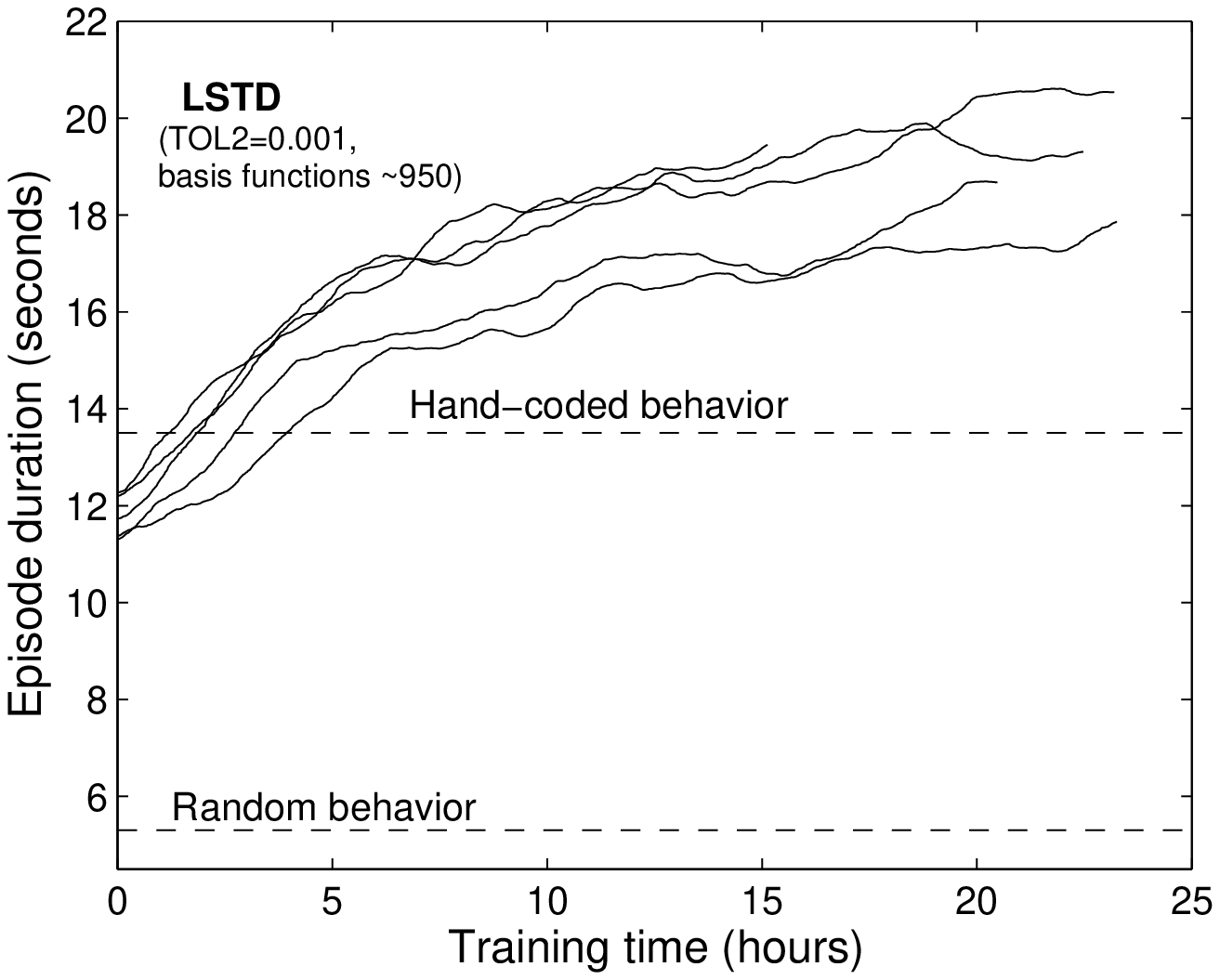}
\includegraphics[width=0.47\textwidth]{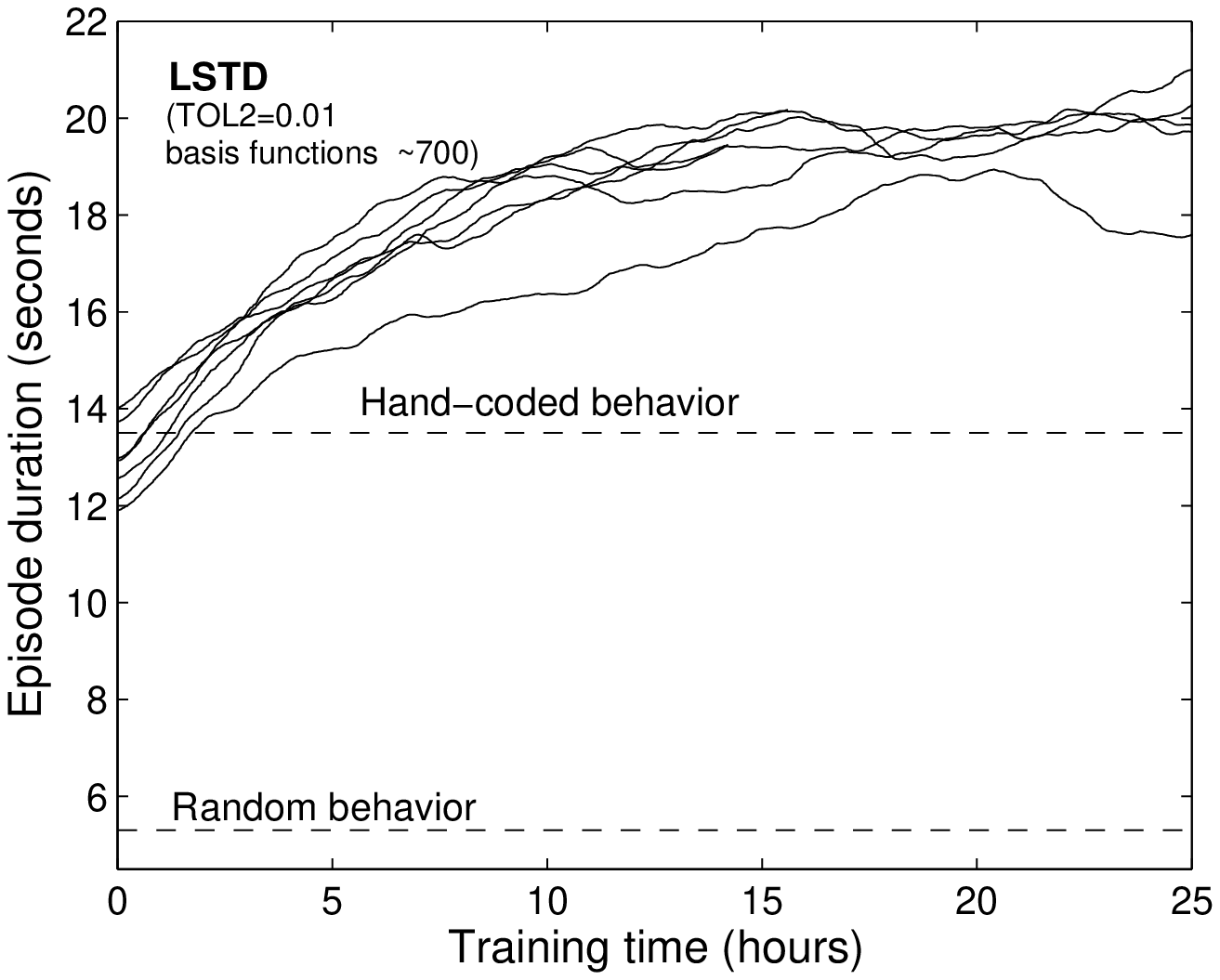}
\includegraphics[width=0.47\textwidth]{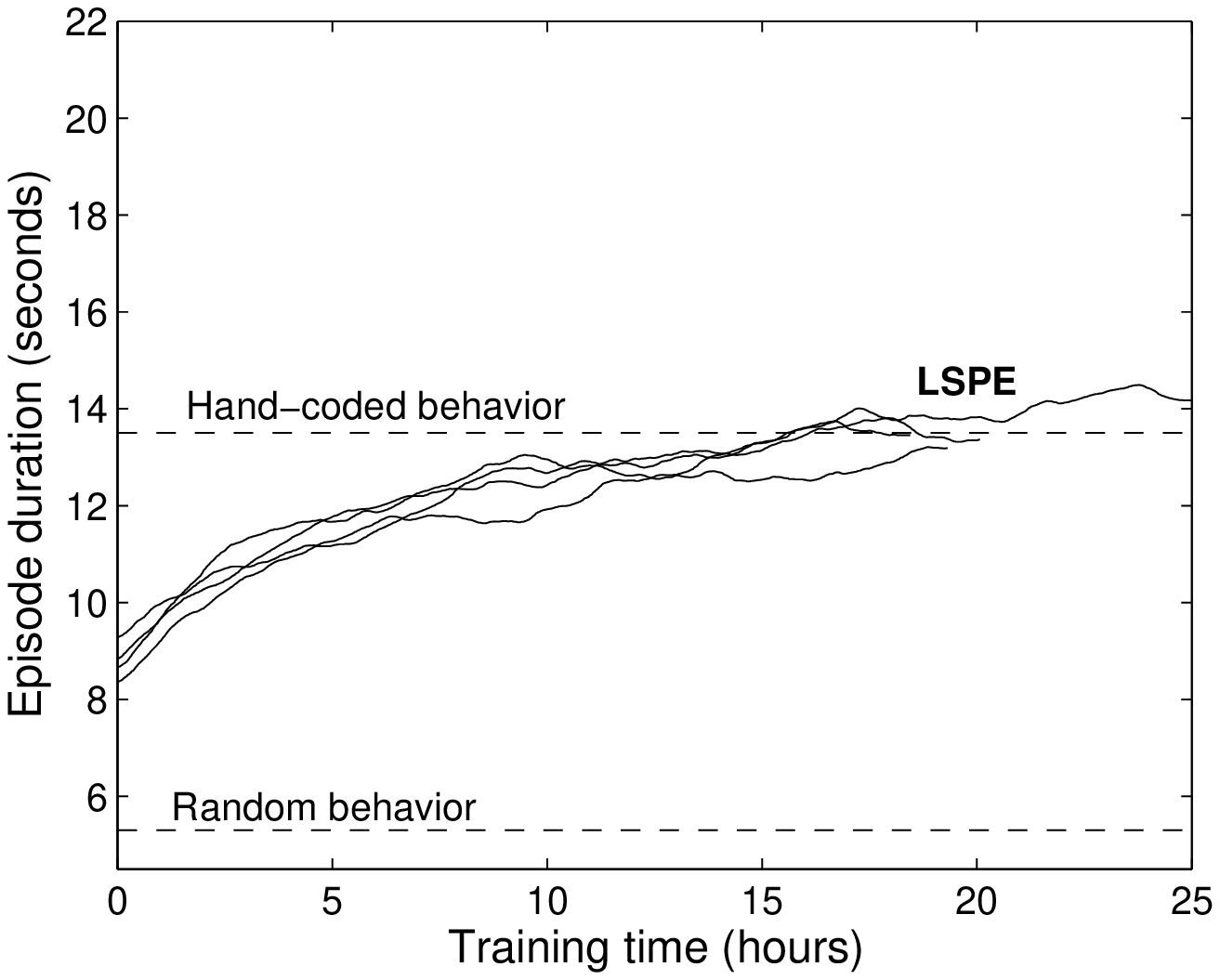}
\includegraphics[width=0.47\textwidth]{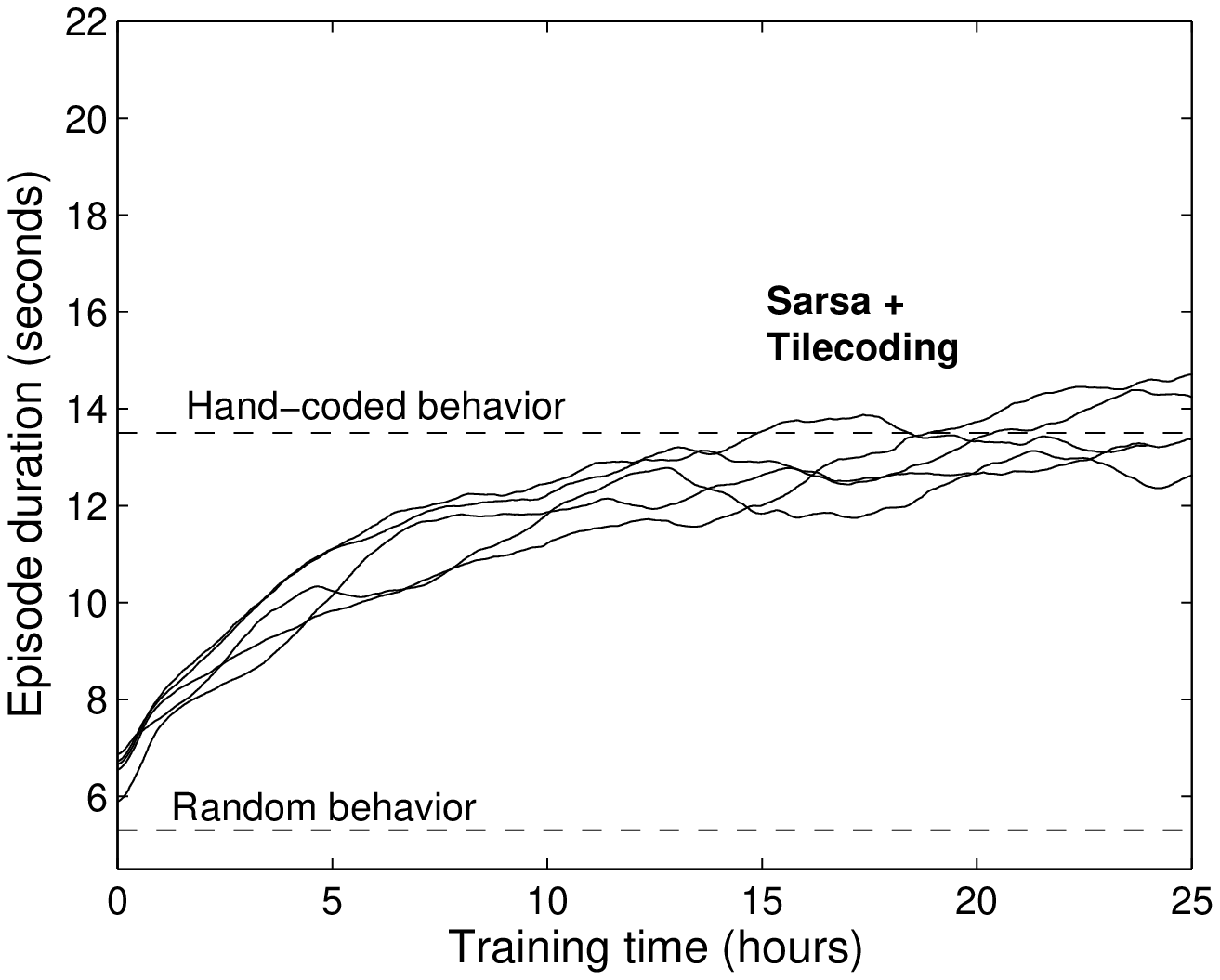}
	\caption{From left to right: Learning curves for our approach with LSTD ({\tt TOL2}=0), LSTD ({\tt TOL2}=0.001),
	LSTD ({\tt TOL2}=0.01), and LSPE. At the bottom we show the curves for Sarsa with tilecoding corresponding to \citep{AB05}. 
	We plot the average time the keepers
	are able to control the ball (quality of learned behavior) against the 	
	training time. After interacting for
	15 hours the performance does not increase any more and the agent has experienced roughly 35,000 state 
	transitions. }
	\label{fig:results}
\end{figure}

\paragraph{Results.}
We evaluate every algorithm/parameter configuration using 5 independent runs. The learning curves for these runs are shown in 
Figure~\ref{fig:results}. The curves plot the average time the keepers are able to keep the ball 
(corresponding to the performance) against the
simulated time the keepers were learning (roughly corresponding to the observed 
training examples). Additionally, two horizontal lines indicate the scores for the two 
benchmark policies random behavior and optimized 
hand-coded behavior used in \citep{AB05}.

The plots show that generally RL is able to learn policies that are at least as effective as the optimized hand-coded
behavior. This is indeed quite an achievement, considering that the latter is the product of considerable manual effort. Comparing
the three approaches Sarsa, LSPE and LSTD we find that the performance of LSPE is on par with Sarsa. The curves of LSTD tell a 
different story however; here we are  outperforming Sarsa by 25\% in terms of performance (in Sarsa the best 
performance is about $15$ seconds,
in LSTD the best performance is about $20$ seconds). This gain is even 
more impressive when we consider the time scale at which this behavior is learned; just after a mere 2 hours we are already 
outperforming hand-coded control. Thus our approach needs far fewer state transitions to discover good behavior. 
The third observation shows the effectiveness of our proposed
supervised basis function selection; here we show that our supervised approach performs as well as the unsupervised one, but requires
significantly fewer basis functions to achieve that level of performance ($\sim$ 700 basis functions at {\tt TOL2}$=0.01$ 
against 1400 basis functions at {\tt TOL2}$=0$).  

Regarding the unexpectedly weak performance of LSPE in comparison with LSTD, we conjecture that this strongly depends on the underlying 
architecture of policy iteration (i.e.\ OPI vs. actor-critic) as well as the specific learning problem. On a related number of 
experiments carried out with the octopus arm benchmark\footnote{From the ICML06 RL benchmarking page: \newline {\tt http://www.cs.mcgill.ca/dprecup/workshops/ICML06/octopus.html}}
we made exactly the opposite observation \citep[not discussed here in more detail, see][]{mein_ieee_adprl2007}.

\section{Discussion and related work}
\vspace*{-0.125cm}
We have presented a kernel-based approach for least-squares based policy evaluation in RL using regularization networks
as underlying function approximator. The key point is an efficient supervised basis selection mechanism, which is used to 
select a subset of relevant basis functions directly from the data stream.  
The proposed method was particularly devised with high-dimensional, stochastic control tasks for RL in mind; 
we prove its effectiveness using the RoboCup keepaway benchmark. Overall the results indicate that kernel-based online learning in 
RL is very well possible and recommendable. Even the rather few simulation runs we made clearly show that our approach is superior to convential 
function approximation in RL using grid-based tilecoding. What could be even more important is that the kernel-based approach
only requires the setting of some fairly general parameters that do not depend on the specific control problem one wants to solve. 
On the other hand, using tilecoding or a fixed basis function network in high dimensions requires considerable manual effort on part of 
the programmer to carefully devise problem-specific features and manually choose suitable basis functions.

\citet{engel2003gptd,engel2005rlgptd} initially advocated using kernel-based methods in RL and proposed the 
related GPTD algorithm. Our method using regularization
networks develops this idea further. Both methods have in common the online selection of relevant 
basis functions based on \citep{csato2001sparse}. As opposed to the unsupervised selection in GPTD, 
we use a supervised criterion to further reduce the number of relevant basis functions selected.
A more fundamental difference is the policy 
evaluation method addressed by the respective formulation; GPTD models the Bellman residuals and corresponds to the BRM approach 
(see Section 2.1.2). Thus, in its original formulation GPTD can be only applied to RL problems with deterministic state transitions.
In contrast, we provide a unified and concise formulation of LSTD and LSPE which can deal with stochastic state transitions as well.  
Another difference is the type of benchmark problem used to showcase the respective method;         
GPTD was demonstrated by learning to control a simulated octopus arm, which was posed as an 88-dimensional control 
problem \citep{engel2005octopus}. Controlling the octopus arm is a deterministic control problem  with known state transitions 
and was solved there using model-based RL. In contrast, 3vs2 keepaway is only a 13-dimensional problem; here however, 
we have to deal with 
stochastic and unknown state transitions and need to use model-free RL.

\acks
The authors wish to thank the anonymous reviewers for their useful comments and suggestions.

\begin{appendix}

\section{A summary of the updates}
%
Let $\bxtt=(s_{t+1},a_{t+1})$ be the next state-action tuple and $r_{t+1}$ be the reward assiociated 
with transition from the previous state $s_t$ to $s_{t+1}$ under $a_t$. Define the abbreviations:
\begin{flalign*}
\bkt  &\definedas \bkm{\bx_t}     &    \bktt &\definedas \bkm{\bxtt}   &     \bhtt&\definedas \bkt - \gamma \bktt\\
k^*_{t}&\definedas k(\bxt,\bxtt)  & k^*_{t+1}&\definedas k(\bxtt,\bxtt)& h^*_{t+1}&\definedas k^*_{t}-\gamma k^*_{t+1}  
\end{flalign*}
and $\batt \definedas \bKmmi \bktt$.

\subsection{Unsupervised basis selection}
We want to test if $\bxtt$ is well represented by the current basis functions in the dictionary 
or if we need to add $\bxtt$ to the basis elements. Compute 
\[
    \delta=\kstar_{t+1} - \bktt\trans \batt. \tag{\ref{eq:ALD-test}}
\]
If $\delta< \texttt{TOL1}$, then add $\bxtt$ to the dictionary, execute the growing step (see below) and update
\[
\bK_{m+1,m+1}^{-1}=\begin{bmatrix} \bKmmi & \bzeros \\ \bzeros\trans & 0\end{bmatrix}
+\frac{1}{\delta}\begin{bmatrix} -\batt \\ 1 \end{bmatrix}
\begin{bmatrix} -\batt \\ 1 \end{bmatrix}\trans. \tag{\ref{eq: aktualisiere Kmm}}
\]

\subsection{Recursive updates for BRM}
\begin{itemize} 
       \item Normal step $\{t,m\} \mapsto \{t+1,m\}$: 
       \begin{enumerate}
           \item \[ \bPittm =  \bPitm - \frac{\bPitm\bhtt \bhtt\trans\bPitm}{\Delta} 
                         \tag{\ref{eq:normal Pittm}} \]
                     with $\Delta=1+\bhtt\trans \bPitm \bhtt$.
           \item \[ \bwttm  =  \bwtm + \frac{\varrho}{\Delta} \bPitm \bhtt
                         \tag{\ref{eq:normal wttm}} \]  
                     with $\varrho=\rtt - \bhtt\trans \bwtm$.
       \end{enumerate}
       \item Growing step $\{t+1,m\} \mapsto \{t+1,m+1\}$
       \begin{enumerate}
            \item \[ \bPittmm=\begin{bmatrix} \bPittm & \bzeros \\ \bzeros & 0 \end{bmatrix} + 
                        \frac{1}{\Delta_b} \begin{bmatrix} -\bwb \\ 1 \end{bmatrix}
                          \begin{bmatrix} -\bwb \\ 1 \end{bmatrix}\trans 
                          \tag{\ref{eq:Pitmm}}
                          \]
                     where 
                     \[                    
                         \bwb =\batt + \frac{\delta_h}{\Delta} \bPitm \bhtt, \qquad
			                   \Delta_b = \frac{\delta_h^2}{\Delta}+\sigma^2\delta_h, \qquad
			                   \delta_h=\hstar_{t+1} - \bhtt\trans \batt
                     \]
             \item \[
                       \bwttmm=\begin{bmatrix} \bwttm \\ 0 \end{bmatrix} + \kappa 
                        \begin{bmatrix} -\bwb \\ 1 \end{bmatrix}
                        \tag{\ref{eq:bbetatmm}}
                   \]  
                   where $\kappa=-\frac{\delta_h\varrho}{\Delta_b\Delta}$.      
       \end{enumerate}
       \item{Reduction of regularized cost when adding $\bxtt$ (supervised basis selection)}:
              \[  \xittmm=\xittm- \kappa^2 \Delta_b \tag{\ref{eq:xitmm}} \]
             For supervised basis selection we additionally check if 
             $\kappa^2 \Delta_b > \texttt{TOL2}$.
\end{itemize}

\subsection{Recursive updates for LSTD($\lambda$)}
\newcommand{\bwba}{\bwb^{(1)}}
\newcommand{\bwbb}{\bwb^{(2)}}
\newcommand{\da}{\delta^{(1)}}
\newcommand{\db}{\delta^{(2)}}
\begin{itemize} 
       \item Normal step $\{t,m\} \mapsto \{t+1,m\}$: 
       \begin{enumerate}
            \item \[ \bzttm=(\gamma\lambda)\bztm+\bkt \]
           \item \[ \bPittm =  \bPitm - \frac{\bPitm\bzttm \bhtt\trans\bPitm}{\Delta}
                        \tag{\ref{eq:normal Pittm}'} \]
                     with $\Delta=1+\bhtt\trans \bPitm \bzttm$. 
           \item \[ \bwttm  =  \bwtm + \frac{\varrho}{\Delta} \bPitm \bzttm
                        \tag{\ref{eq:normal wttm}'} \]  
                     with $\varrho=\rtt - \bhtt\trans \bwtm$.
       \end{enumerate}
       \item Growing step $\{t+1,m\} \mapsto \{t+1,m+1\}$
       \begin{enumerate}
            \item \[
                       \bzttmm=\begin{bmatrix} \bzttm\trans & z_{t+1,m}^* \end{bmatrix}\trans 
                   \]
                   where $z_{t+1,m}^*=(\gamma\lambda)\bztm\trans\batt + k^*_t$.  
            \item \[ \bPittmm=\begin{bmatrix} \bPittm & \bzeros \\ \bzeros & 0 \end{bmatrix} + 
                        \frac{1}{\Delta_b} \begin{bmatrix} -\bwba \\ 1 \end{bmatrix}
                          \begin{bmatrix} -\bwbb & 1 \end{bmatrix}
                          \tag{\ref{eq:Pitmm}'} \]
                     where 
                     \begin{flalign*}
                        \bwba & =\batt + \frac{\da}{\Delta} \bPitm \bzttm & 
                        \da &=h^*_{t+1} - \batt\trans\bhtt \\
                        \bwbb & =\batt\trans + \frac{\db}{\Delta} \bhtt\trans\bPitm & 
                        \db &=z_{t+1,m}^* - \batt\trans\bzttm \\
                     \end{flalign*}
                     and $\Delta_b=\frac{\da \db}{\Delta} + \sigma^2 (k^*_{t+1}-\bktt\trans\batt)$.
             \item \[
                      \bwttmm=\begin{bmatrix} \bwttm \\ 0 \end{bmatrix} + \kappa 
                        \begin{bmatrix} -\bwba \\ 1 \end{bmatrix} \tag{\ref{eq:bbetatmm}'}
                   \]  
                   where $\kappa=-\frac{\db\varrho}{\Delta_b\Delta}$.

       \end{enumerate}
\end{itemize}

\subsection{Recursive updates for LSPE($\lambda$)}

\begin{itemize} 
       \item Normal step $\{t,m\} \mapsto \{t+1,m\}$: 
       \begin{enumerate}
           \item \begin{eqnarray*}
                      \bzttm &=&(\gamma\lambda)\bztm+\bktt \\
                      \bAttm &=& \bAtm+\bzttm\bhtt\trans \\
                      \bbttm &=& \bbtm+\bzttm r_{t+1}                      
                  \end{eqnarray*}
           \item \[ \bPittm =  \bPitm - \frac{\bPitm\bktt \bktt\trans\bPitm}{\Delta} 
                    \tag{\ref{eq:normal Pittm}''} \]
                     with $\Delta=1+\bktt\trans \bPitm \bktt$. 
           \item \[ \bwttm  =  \bwtm + \eta \bPittm (\bbttm - \bAttm \bwtm)
                     \tag{\ref{eq:normal wttm}''} \]  
       \end{enumerate}
       \item Growing step $\{t+1,m\} \mapsto \{t+1,m+1\}$
       \begin{enumerate}
            \item \begin{gather*}
                       \bzttmm=\begin{bmatrix} \bzttm \\ z_{t+1,m}^* \end{bmatrix} 
                       \qquad \qquad \bbttmm=\begin{bmatrix} 
                                \bbttm \\ \batt\trans\bbtm + z_{t+1,m}^* r_{t+1}
                             \end{bmatrix} \\
                       \bAttmm=\begin{bmatrix} 
                              \bAttm        & \bAtm\batt+\bzttm h^* \\
                              \batt\trans\bAtm+z_{t+1,m}^*\bhtt\trans  &\batt\trans\bAtm\batt+ z_{t+1,m}^* h^*
                          \end{bmatrix}                      
                   \end{gather*}
                   where $z_{t+1,m}^*=(\gamma\lambda)\bztm\trans\batt + k^*_t$.                     
            \item \[ \bPittmm=\begin{bmatrix} \bPittm & \bzeros \\ \bzeros & 0 \end{bmatrix} + 
                        \frac{1}{\Delta_b} \begin{bmatrix} -\bwb \\ 1 \end{bmatrix}
                          \begin{bmatrix} -\bwb \\ 1 \end{bmatrix}\trans
                          \tag{\ref{eq:Pitmm}''}
                          \]
                     where 
                     \[                    
                         \bwb =\batt + \frac{\delta}{\Delta} \bPitm \bktt, \qquad
			                   \Delta_b = \frac{\delta^2}{\Delta}+\sigma^2\delta, \qquad
			                   \delta=\kstar_{t} - \bkt\trans \batt
                     \]

                     and $\Delta_b=\frac{\da \db}{\Delta} + \sigma^2 (k^*_{t+1}-\bktt\trans\batt)$.
             \item \[
                      \bwttmm=\begin{bmatrix} \bwttm \\ 0 \end{bmatrix} + \kappa 
                        \begin{bmatrix} -\bwba \\ 1 \end{bmatrix}\tag{\ref{eq:bbetatmm}''}
                   \]  
                   where $\kappa=-\frac{\db\varrho}{\Delta_b\Delta}$.                         
       \end{enumerate}
      \item{Reduction of regularized cost when adding $\bxtt$ (supervised basis selection)}:
         \[  \xittmm=\xittm - \Delta_b^{-1}(c-\bwb\trans \bd)^2  \tag{\ref{eq:xitmm}''} \]
           where $c=\batt\trans(\bbtm-\bAtm\bwtm)+z^*_{t+1,m}(\rtt-\bhtt\trans\bwtm)$ and 
           $\bd=\bbttm - \bAttm\bwtm$. For supervised basis selection we additionally check if 
             $\Delta_b^{-1}(c-\bwb\trans \bd)^2 > \texttt{TOL2}$.
                     
\end{itemize}

\end{appendix}

\bibliographystyle{plainnat}

\end{document}